\documentclass[journal]{IEEEtran}
\usepackage{amsmath}
\usepackage{graphicx}
\usepackage{threeparttable}
\usepackage{multirow}
\usepackage{colortbl}

\usepackage{caption}
\usepackage{subfig}

\ifCLASSINFOpdf
\else
\fi
\begin{document}

\title{Constrained Deep Weak Supervision \\for Histopathology Image Segmentation}
%
%
%

\author{Zhipeng Jia, Xingyi Huang, Eric I-Chao Chang and Yan Xu*
\thanks{This work is supported by Microsoft Research under the eHealth program, the Beijing National Science Foundation in China under Grant 4152033, the Technology and Innovation Commission of Shenzhen in China under Grant shenfagai2016-627, Beijing Young Talent Project in China, the Fundamental Research Funds for the Central Universities of China under Grant SKLSDE-2015ZX-27 from the State Key Laboratory of Software Development Environment in Beihang University in China. \emph{Asterisk indicates corresponding author.}}
\thanks{Xingyi Huang, and Yan Xu are with State Key Laboratory of Software Development Environment and Key Laboratory of Biomechanics and Mechanobiology of Ministry of Education and Research Institute of Beihang University in Shenzhen, Beihang University, Beijing 100191, China (email: huangxingyi102@126.com; xuyan04@gmail.com).}
\thanks{Zhipeng Jia, Eric I-Chao Chang, and Yan Xu are with Microsoft Research, Beijing 100080, China (email: zhipeng.jia@outlook.com; echang@microsoft.com; xuyan04@gmail.com).}
\thanks{Zhipeng Jia is with Institute for Interdisciplinary Information Sciences, Tsinghua University, Beijing 100084, China (email: zhipeng.jia@outlook.com).}
}

\maketitle

\begin{abstract}
In this paper, we develop a new weakly-supervised learning algorithm to learn to segment cancerous regions in histopathology images.
Our work is under a multiple instance learning framework (MIL) with a new formulation, deep weak supervision (DWS); we also propose an effective way to introduce constraints to our neural networks to assist the learning process.
The contributions of our algorithm are threefold:
(1) We build an end-to-end learning system that segments cancerous regions with fully convolutional networks (FCN) in which image-to-image weakly-supervised learning is performed.
(2) We develop a deep week supervision formulation to exploit multi-scale learning under weak supervision within fully convolutional networks.
(3) Constraints about positive instances are introduced in our approach to effectively explore additional weakly-supervised information that is easy to obtain and enjoys a significant boost to the learning process.
The proposed algorithm, abbreviated as DWS-MIL, is easy to implement and can be trained efficiently. Our system demonstrates state-of-the-art results on large-scale histopathology image datasets and can be applied to various applications in medical imaging beyond histopathology images such as MRI, CT, and ultrasound images.
\end{abstract}


\IEEEpeerreviewmaketitle

\section{Introduction}
\IEEEPARstart
{H}{igh} resolution histopathology images play a critical role in cancer diagnosis, providing essential information to separate non-cancerous tissues from cancerous ones. A variety of classification and segmentation algorithms have been developed in the past \cite{Esgiar02, Huang, Madabhushi_2009, Park, Tabesh, madabhushi2016image,gurcan2009histopathological,tang2009computer}, focusing primarily on the design of local pathological patterns, such as morphological \cite{Huang}, geometric \cite{Esgiar02}, and texture \cite{Kong2009} features  based on various clinical characteristics.

In medical imaging, supervised learning approaches \cite{zheng2008four,tu2008auto,criminisi2013decision} have shown their particular effectiveness in performing image classification and segmentation for modalities such as MRI, CT, and Ultrasound. However, the success of these supervised learning algorithms depends on the availability of a large amount of high-quality manual annotations/labeling that are often time-consuming and costly to obtain. In addition, well-experienced medical experts themselves may have a disagreement on ambiguous and challenging cases. Unsupervised learning strategies where no expert annotations are needed point to a promising but thus far not clinically practical direction.

In-between supervised and unsupervised learning, weakly-supervised learning in which only coarse-grained (image-level) labeling is required makes a good balance of
having a moderate level of annotations by experts while
being able to automatically explore fine-grained (pixel-level) classification \cite{dietterich1997solving, zhang2005multiple, Maron:icml98, xu2014weakly, xu2014deep, pathak2015constrained, xu2012multiple, xucontexts}. In pathology, a pathologist annotates whether a given histopathology image has a cancer or not; a weakly-supervised learning algorithm would hope to automatically detect and segment cancerous tissues based on a collection of histopathology (training) images annotated by expert pathologists;
this process that substantially reduces the amount of work for annotating cancerous tissues/regions
falls into the category of weakly-supervised learning, or more specifically multiple instance learning \cite{dietterich1997solving}, which is the main topic of this paper.

\begin{figure}[!th]
\centering
\includegraphics[width=90mm]{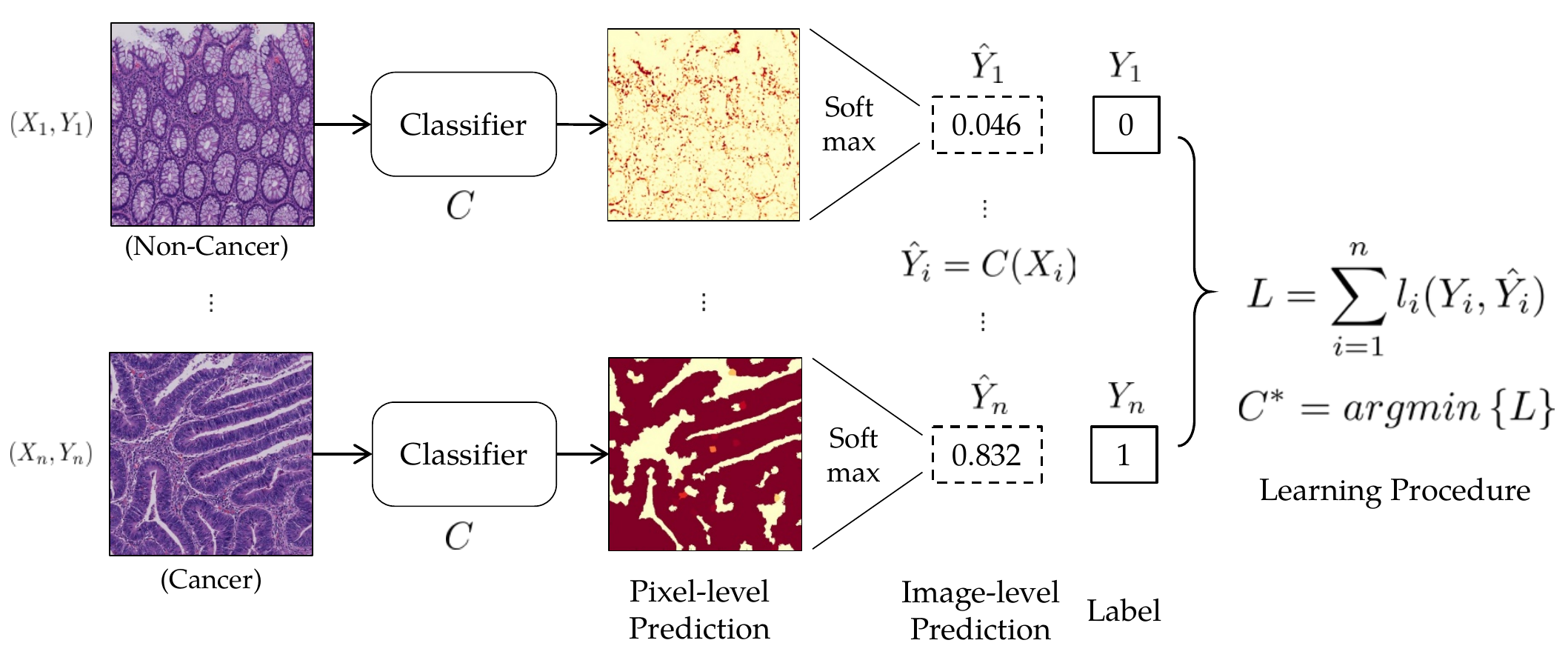}
\caption{Illustration of the learning procedure of a MIL algorithm.
Our training dataset is denoted by $S=\{(X_i, Y_i), i=1, 2, 3, \ldots, n\}$, where $X_i$ indicates the $i$th input image, and $Y_i\in\{0,1\}$ represents its corresponding manual label ($Y_i=0$ refers to a non-cancer image and $Y_i=1$ refers to a cancer image).
Given an input image, a classifier $C$ generates pixel-level predictions.
Then, the image-level prediction $\widehat{Y}_i$ is computed from pixel-level predictions via a softmax function.
Next, a loss between the ground truth $Y_i$ and the image-level prediction $\widehat{Y}_i$ is computed for the $i$th input image, denoted by $l_i(Y_i, \widehat{Y}_i)$.
Finally, an objective loss function $\mathnormal{L}$ takes the sum of loss functions of all input images. The classifier $C$ is learned by minimizing the objective loss function.}
\label{fig:flowchart_MIL}
\end{figure}

Multiple instance learning (MIL) was first introduced by Dietterich et al. \cite{dietterich1997solving} to predict drug activities; a wealthy body of
MIL based algorithms was developed thereafter \cite{Andrews:nips02, maron1998framework, zhang2005multiple}.
In multiple instance learning, instances arrive together in groups during training, known as \textit{bags}, each of which is assigned either a positive or a negative label (can be multi-class), but absent instance-level labels (as shown in Figure \ref{fig:flowchart_MIL}). In the  original MIL setting \cite{dietterich1997solving}, each bag consists of a number of organic molecules as instances; their task was to predict instance-level label for the training/test data, in addition to being able to perform bag-level classification.
In our case here, each histopathology image with cancer or non-cancer label forms a bag and each pixel in the image is referred to as an instance (note that the instance features are computed based on each pixel's surroundings beyond the single pixel itself).

Despite the great success of MIL approaches \cite{dietterich1997solving, zhang2005multiple, Maron:icml98} that explicitly deal with the latent (instance-level) labels, one big problem with many existing MIL algorithms is the use of pre-specified features \cite{Andrews:nips02, zhang2005multiple, xu2014weakly}. Although algorithms like MILBoost \cite{zhang2005multiple} have embedded feature selection procedures, their input feature types are nevertheless fixed and pre-specified. To this point, it is natural to develop an integrated framework by combining the MIL concept with convolutional neural networks (CNN), which automatically learns rich hierarchical features for pattern recognition with state-of-the-art classification/recognition results.
A previous approach that adopts CNN in a MIL formulation was recently proposed \cite{xu2014deep}, but its greatest limitation is the use of image patches instead of full images, making the learning process slow and ineffective. For patch-based approaches: (1) image patch size has to be specified in advance; (2) every pixel as the center of a patch is potentially an instance, resulting in millions of patches to be extracted even for a single image; (3) feature extraction for image patches is not efficient. Beyond the patch-centric CNN framework is the image-centric paradigm where image-to-image prediction can be performed by fully convolutional networks (FCN) \cite{long2015fully} in which features for all pixels are computed altogether. The efficiency and effectiveness of both training and testing by FCN family models have shown great success in various computer vision applications such as image labeling \cite{long2015fully, chen2014semantic} and edge detection \cite{xie2015deep}. An early version of FCN applied in MIL was proposed in \cite{pathak2014fully} which was extended into a more advanced model \cite{pathak2015constrained}.

In this paper, we first build an FCN based multiple instance learning framework to serve as our baseline algorithm for weakly-supervised learning of histopathology image segmentation.
The main focus of this paper is the introduction of deep weak supervision and constraints to our multiple instance learning framework. We abbreviate our deep weak supervision for multiple instance learning as DWS-MIL and our constrained deep weak supervision for multiple instance learning as CDWS-MIL.
The concept of deep supervision in the supervised learning was introduced in \cite{lee2015deeply}, which is combined with FCN for edge detection \cite{xie2015deep}.
We propose a deep weak supervision strategy in which the intermediate FCN layers are expected to be further guided
through weakly-supervised information within their own layers.

We also introduce area constraints that only require a small amount of additional labeling effort but are shown to be immensely effective. That is, in addition to the annotation of being a cancerous or non-cancerous image, we ask pathologists to give a rough estimation of the relative size (e.g $30\%$) of cancerous regions within each image; this rough estimation is then turned into an area constraint in our MIL formulation.
%
Our motivation to introduce area constrains is three-fold. First, having informative but easy to obtain expert annotation can always help the learning process and we are encouraged to seek information beyond being just positive or negative.
There exists a study in cognitive science \cite{2004NBIC} indicating the natural surfacing of the concept of relative size when making a discrete yes-or-no decision.
Second, our DWS-MIL formulation under an image-to-image paradigm allows the additional term of the area constraints to be conveniently carried out through back-propagation, which is nearly impossible to do if a patch-based approach is adopted \cite{xu2014weakly,xu2014deep}. Third, having area constraints conceptually and mathematically greatly enhances learning
capability; this is evident in our experiments where a significant performance boost is observed using the area constraints.

To summarize, in this paper we develop a new multiple instance learning algorithm for histopathology image segmentation under a deep weak supervision formulation, abbreviated as DWS-MIL.
The contributions of our algorithm include:
(1) DWS-MIL is an end-to-end learning system that performs image-to-image learning and prediction under weak supervision.
(2) Deep weak supervision is adopted in each intermediate layer to exploit nested multi-scale feature learning.
(3) Area constraints are also introduced as weak supervision, which is shown to be particularly effective in the learning process, significantly enhancing segmentation accuracy with very little extra work during the annotation process.
In addition, we experiment with the adoption of super-pixels \cite{achanta2012slic} as an alternative way to pixels and show their effectiveness in maintaining intrinsic tissue boundaries in histopathology images.

\section{Related Work}
\label{related}
Related work can be divided into three broad categories: (1) directly related work, (2) weakly supervised learning in computer vision, and (3) weakly supervised learning in medical images.

\subsection{ Directly related work}
Three existing approaches that are closely related to our work are discussed below.

Xu et al. \cite{xu2014weakly} propose a histopathology image segmentation algorithm in which the concept of multiple clustered instance learning (MCIL) is introduced.
The MCIL algorithm \cite{xu2014weakly} can simultaneously perform image-level classification, patch-level segmentation and patch-level clustering.
However, as mentioned previously, their approach is a patch-based system that is extremely space-demanding (requiring large disk space to store the features) and time-consuming to train. In addition, a boosting algorithm is adopted in \cite{xu2014weakly} with all feature types pre-specified, but features in our approaches are automatically learned.

Pathak et al. present an early version of fully convolutional networks applied in a multiple instance learning setting \cite{pathak2014fully} and they later generalize the algorithm by introducing a new loss function to optimize for any set of linear constraints on the output space \cite{pathak2015constrained}. Some typical linear constraints include suppression, foreground, background, and size constraints. Compared with the generalized constrained optimization in their model, the area constraints proposed in this paper are simpler to carry out through back-propagation within MIL. Moreover, our formulation of deep weak supervision combined with area constraints demonstrates its particular advantage in histopathology image segmentation where only two-class (positive and negative) classification is studied.

Holistically-nested edge detector (HED) is developed in \cite{xie15hed} by combining deep supervision with fully convolutional networks to effectively learn edges and object boundaries. Our deep weak supervision formulation is inspired by HED but we instead focus on a weakly-supervised learning setting as opposed to being fully supervised in HED. Our deep weak supervision demonstrates its power under an end-to-end MIL framework.


\subsection{Weakly supervised learning in computer vision}
A rich body of weakly-supervised learning algorithms exists in computer vision and we discuss them in two groupings: segmentation based and detection based.

\textbf{Segmentation.} In computer vision, MIL has been applied to segmentation in many previous systems \cite{VisualConcepts, Wang:ICML13,pinheiro2015image,papandreou2015weakly}. A patch-based approach would extract pre-specified image features from selected image patches \cite{VisualConcepts, Wang:ICML13} and try to learn the hidden instance labeling under MIL.
The limitations of these approaches are apparent, as stated before, requiring significant space and computation.
More recently, convolutional neural networks have become increasingly popular.
Pinheiro et al. \cite{pinheiro2015image} propose a convolutional neural network-based model which weights important pixels during training.
Papandreous et al. \cite{papandreou2015weakly}  propose an expectation-maximization (EM) method using image-level and bounding box annotation in a weakly-supervised setting. 

\textbf{Object detection.} MIL has also been applied to objection detection where the instances are now image patches of varying sizes, which are also referred to as sliding windows. The space for storing all instances are enormous and proposals are often used to limit the number of possible instances \cite{zhu2015unsupervised}. A lot of algorithms exist in this domain and we name a couple here.
Cinbis et al. \cite{cinbis2016weakly} propose a multi-fold multiple instance learning procedure, which prevents training from prematurely looking at all object locations; this method iteratively trains a detector and infers object locations.  
Diba et al. \cite{diba2016weakly} propose a cascaded network structure which is composed of two or three stages and is trained in an end-to-end pipeline.

\subsection {Weakly supervised learning in medical imaging }
Weakly-supervised learning has been applied to medical images as well.
Yan et al. \cite{yan2016multi} propose a multi-instance deep learning method by automatically discovering discriminative local anatomies for anatomical structure recognition; positive instances are defined as contiguous bounding boxes and negative instances (non-informative anatomy) are randomly selected from the background.
A weakly-supervised learning approach is also adopted in Hou et al. \cite{hou2015efficient} to train convolutional neural networks to identify gigapixel resolution histopathology images.

Though being promising, existing methods in medical imaging lack an end-to-end learning strategy for image-to-image learning and prediction under MIL.

\section{Method}
\label{method}
In this section, we present in detail the concept and formulation of our algorithms.
First, we introduce our baseline algorithm, a method in spirit similar to the FCN-MIL method \cite{pathak2014fully} but our method focuses on two-class classification whereas FCN-MIL is a multi-class approach with some preliminary results shown for natural image segmentation.
We then discuss the main part of this work, deep weak supervision for MIL (DWS-MIL) and constrained deep weak supervision for MIL (CDWS-MIL).

\subsection{Our Baseline}
\label{baseline}
Here, we build an end-to-end MIL method as our baseline to perform image-to-image learning and prediction, in which the MIL formulation enables automatic learning of pixel-level segmentation from image-level labels.

We denote our training dataset by $S=\{(X_i, Y_i), i=1, 2, 3, \ldots, n\}$, where $X_i$ denotes the $i$th input image and $Y_i\in\{0,1\}$ refers to the manual annotation (ground truth label) assigned to the $i$th input image.
Here $Y_i=0$ refers to a non-cancer image and $Y_i=1$ refers to a cancerous image.
Figure \ref{fig:flowchart_MIL} demonstrates the basic concept.
As mentioned previously, our task is to be able to perform pixel-level prediction learned from image-level labels and each pixel is referred to as an instance in this case.
We denote $\widehat{Y}_{ik}$ to be the probability of the $k$th pixel being positive in the $i$th image, where $k=\{1, 2, \ldots, |X_i|\}$ and $|X_i|$ represents the total number of pixels of image $X_i$.
If an image-level predictions $\widehat{Y}_i$ can be computed from all $\widehat{Y}_{ik}$s, then it can be used against the true image-level labels $Y_i$ to calculate a loss $\mathcal{L}_{mil}$.
The loss function we opt to use is the cross-entropy cost function:
\begin{equation} \mathcal{L}_{mil}=\sum_i \left(\boldsymbol{I} (Y_i=1)\log \widehat{Y}_i + \boldsymbol{I} (Y_i=0)\log(1-\widehat{Y}_i)\right), \end{equation}
where $\boldsymbol{I}(\cdot)$ is an indicator function.

Since one image is identified to be negative if and only if there does not exist any positive instances, $\widehat{Y}_i$ is typically obtained by $\widehat{Y}_i=\max_k \widehat{Y}_{ik}$, resulting in a \textit{hard maximum} approach.
However, there are two problems with the hard maximum approach: (1) It makes the derivative ${\partial \widehat{Y}_i}/{\partial \widehat{Y}_{ik}}$ discontinuous, leading to numerical instability; (2) ${\partial \widehat{Y}_i}/{\partial \widehat{Y}_{ik}}$ would be 0 for all but the maximum $\widehat{Y}_{ik}$, rendering the learner unable to consider all instances simultaneously.
Therefore, a softmax function is often used to replace the hard maximum approach. We use \textit{Generalized Mean (GM)} as our softmax function \cite{zhang2005multiple}, which is defined as
\begin{equation} \widehat{Y}_i=\left(\frac{1}{|X_i|}\sum_{k=1}^{|X_i|} \widehat{Y}_{ik}^r\right)^{1/r}. \end{equation}
The parameter $r$ controls the sharpness and proximity to the hard function: $\widehat{Y}_i\to\max_k \widehat{Y}_{ik}$ as $r\to\infty$.

We replace classifier $C$ in Figure \ref{fig:flowchart_MIL} with a fully convolutional network (FCN) \cite{long2015fully} using a trimmed VGGNet \cite{simonyan2014very} under the MIL setting. To minimize the loss function via back propagation, we calculate ${\partial \mathcal{L}_{mil}}/{\partial \widehat{Y}_{ik}}$ from ${\partial \mathcal{L}_{mil}}/{\partial \widehat{Y}_i}$. By the chain rule of differentiation,
\begin{equation}
\frac{\partial \mathcal{L}_{mil}}{\partial \widehat{Y}_{ik}}=\frac{\partial \mathcal{L}_{mil}}{\partial \widehat{Y}_i}\frac{\partial \widehat{Y}_i}{\partial \widehat{Y}_{ik}}.
\end{equation}
It suffices to know ${\partial \widehat{Y}_i}/{\partial \widehat{Y}_{ik}}$, whose analytical expression can be derived from the softmax function itself. Once ${\partial \mathcal{L}_{mil}}/{\partial \widehat{Y}_{ik}}$ is known, back propagation can be performed. 

\begin{figure}[!ht]
\centering
\subfloat[training image]{\includegraphics[height=0.28\linewidth]{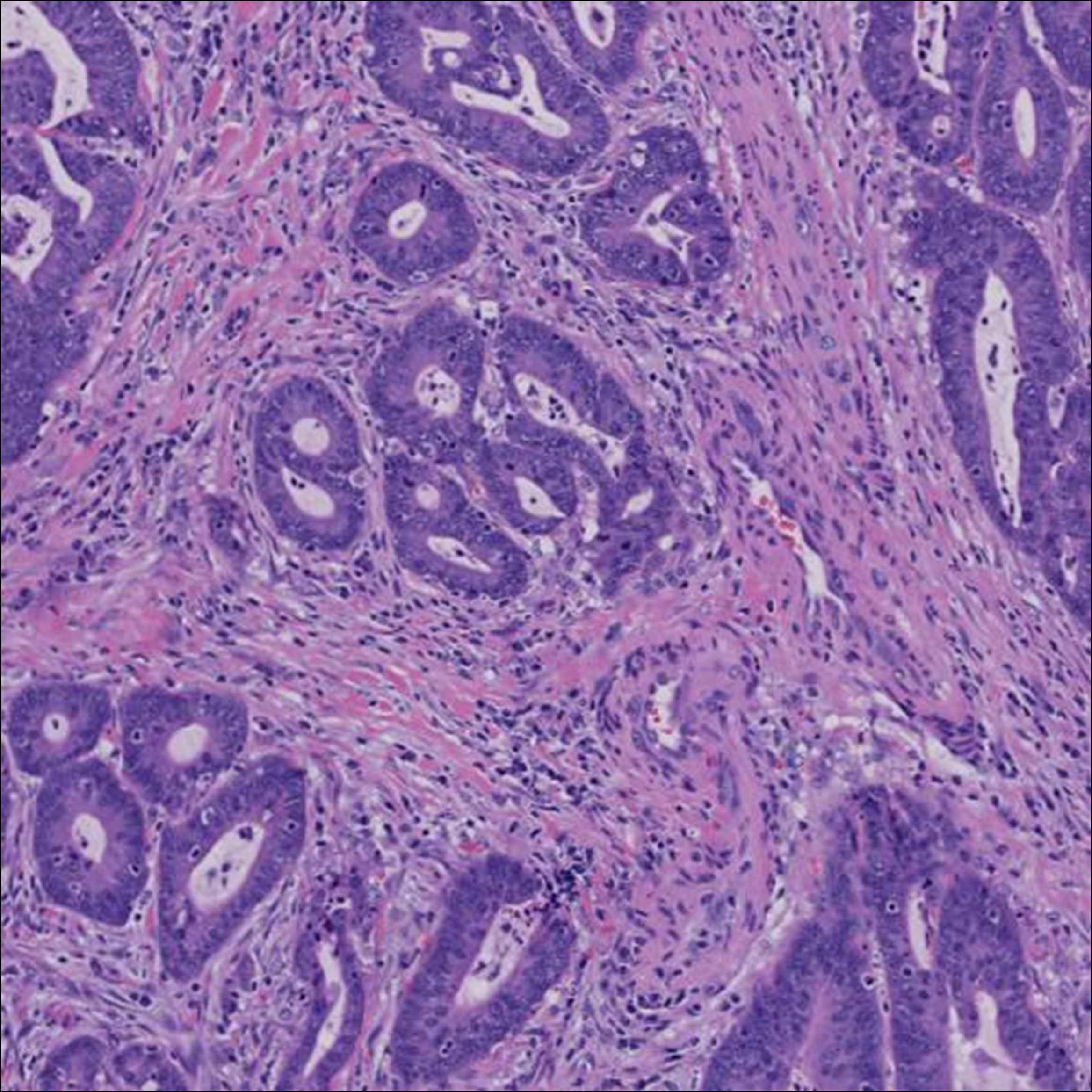}}
\hspace{0.02\linewidth}
\subfloat[heatmap]{\includegraphics[height=0.28\linewidth]{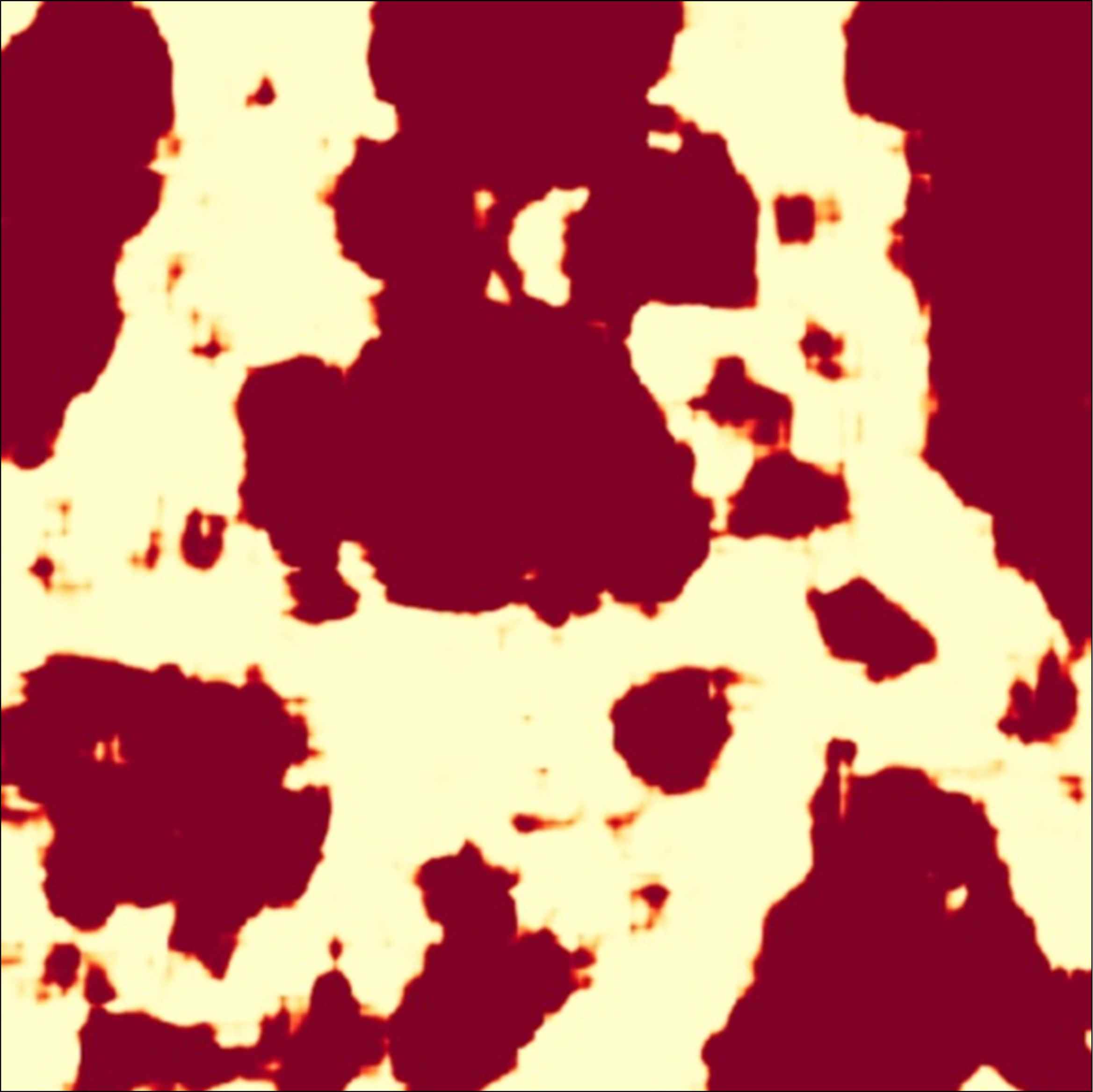}}
\hspace{0.008\linewidth}
\subfloat{\includegraphics[height=0.28\linewidth]{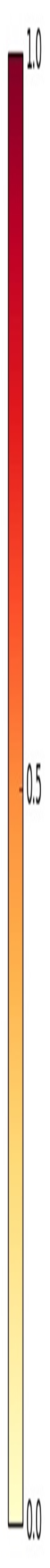}}
\hspace{0.008\linewidth}
\caption{Probability map of an image for all instances. (a) Training image. (b) Instance-level probabilities (segmentation) of being positive (cancerous) by our baseline algorithm. The color coding bar indicates a probability ranging between 0 and 1.}
\label{fig:sample_heatmap}
\end{figure}

In Figure \ref{fig:sample_heatmap}, a training image and its learned instance-level predictions are illustrated. Instance-level predictions are shown as a heatmap,
which shows the probability of each pixel being cancerous.
We use a color coding bar to illustrate the probabilities ranging between 0 and 1.
Note that in the following figures, the instance-level predictions (segmentation) are all displayed as heatmaps and we no longer show the color coding bar for simplicity.

\begin{figure*}[!t]
\centering
\includegraphics[width=150mm]{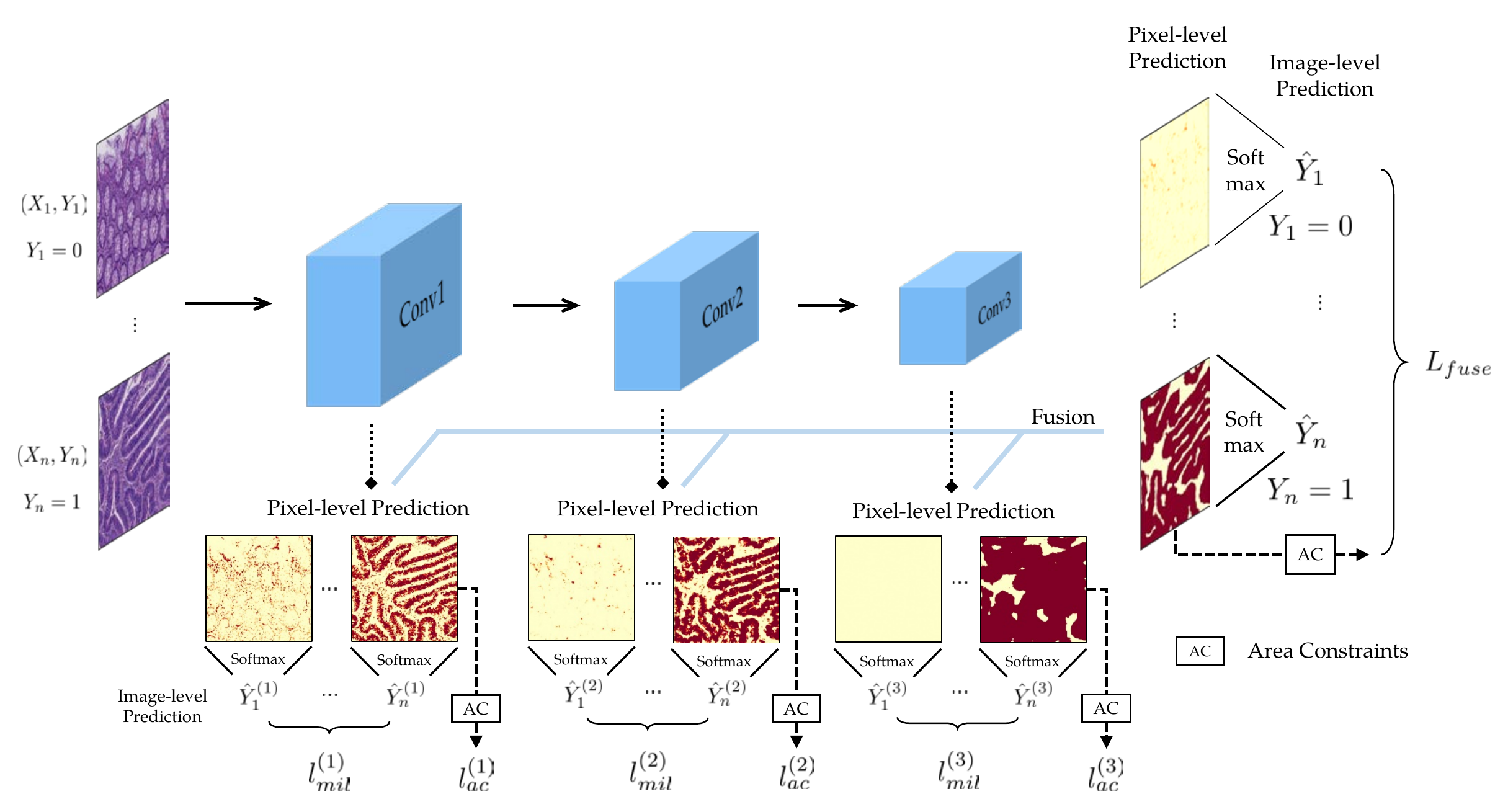}
\caption{Overview of our framework. Under the MIL setting, we adopt first three stages of the VGGNet and connect side-output layers with deep weak supervision under MIL. We also propose area constraints to regularize the size of predicted positive instances. To utilize the multi-scale predictions of the individual layers, we merge side-outputs via a weighted fusion layer. The overall model of equation \eqref{eq:overall} is trained via back-propagation using the stochastic gradient descent algorithm.  }
\label{fig:flowchart_DWS}
\end{figure*}

\subsection{ Constrained Deep Weak Supervision.}
\label{formulation}
After the introduction of our baseline algorithm that is an FCN-like model under MIL, we are ready to introduce the main part of our algorithm, constrained deep weak supervision for histopathology image segmentation.

We denote our training set as $S=\{(X_i, Y_i, a_i), i=1, 2, 3, \ldots, n\}$, where $X_i$ refers to the $i$th input image,  $Y_i\in\{0,1\}$ indicates the corresponding ground truth label for the $i$th  input image, and $a_i$ specifies a rough estimation of the relative area size of the cancerous region within image $X_i$. The $k$th pixel in the $i$th image is given a prediction of the probability being positive, denoted as $\widehat{Y}_{ik}$, where $k=\{1, 2, \ldots, |X_i|\}$ and there are $|X_i|$ pixels in the $i$th image. We denote parameters of the network as $\theta$ and the model is trained to minimize a total loss.

\textbf{Deep weak supervision.}
Aiming to control and guide the learning process across multiple scales, we introduce deep weak supervision by producing side-outputs, forming the multiple instance learning framework with deep weak supervision, named DWS-MIL. The concept of side-output is similar to that defined in \cite{xie15hed}.

Suppose there are $T$ side-output layers, then each side-output layer is connected with an accompanying classifier with the weights $w=(w^{(1)}, \ldots,  w^{(T)})$, where $t=\{1, 2, \ldots, T\}$.
Our goal is to train the model by minimizing a loss between output predictions and ground truth, which is described in the form of the cross-entropy loss function $\mathnormal{l}_{mil}^{(t)}$,
indicating the loss produced by the $t$th side-output layer relative to image-level ground truth. The cross-entropy loss function in each side-output layer is defined as
\begin{equation}
\mathnormal{l}_{mil}^{(t)} =  \sum_i \left(\boldsymbol{I} (Y_i=1)\log \widehat{Y}_i^{(t)}+  \boldsymbol{I} (Y_i=0)\log(1-\widehat{Y}_i ^{(t)})\right).
\label{eq:dws_mil_side}
\end{equation}
The loss function brought by the $t$th side-output layer is defined as :
\begin{equation}
\mathnormal{l}_{side}^{(t)}(\theta, w)= \mathnormal{l}_{mil}^{(t)}(\theta, w).
\end{equation}

The objective function is defined as:
\begin{equation}
\label{eq}  \mathcal{L}_{side}(\theta, w)=\sum_{t=1}^T \mathnormal{l}_{side}^{(t)}(\theta, w).                     \end{equation}

\textbf{Deep weak supervision with constraints.}
Our baseline MIL formulation produces a decent result as shown in the experiments but still with room to improve. One problem is that the positive instances predicted by the algorithm tends to progressively outgrow the true cancerous regions. Here we propose to use an area constraint term to constrain the expansion of the positive instances during training and we name our new algorithm as constrained deep weak supervision, abbreviated as CDWS-MIL. 

A rough estimation of the relative size of the cancerous region, $a_i$, is given by the experts during the annotation process. A measure of the overall ``positiveness'' of all the instances in each image is calculated as 
\begin{equation}
v_{i}=\frac{1}{|X_i|}\sum_{k=1}^{|X_i|} \widehat{Y}_{ik}.
\end{equation}
We then define an area constraint as an $L2$ loss:
\begin{equation}
\mathnormal{l}_{ac}= \boldsymbol{I} (Y_i=1) \sum_{i} (v_{i}-a_i)^2.
\label{eq:ac_loss}
\end{equation}
Naturally the loss function for the $t$th side-output layer can be replaced by:
\begin{equation}
\mathnormal{l}_{side}^{(t)}(\theta, w)\leftarrow \mathnormal{l}_{mil}^{(t)}(\theta, w) + \eta_t\cdot\mathnormal{l}_{ac}^{(t)}(\theta, w),
\end{equation}
where $\mathnormal{l}_{mil}^{(t)}(\theta, w)$ denotes the loss function generated in equation \ref{eq:dws_mil_side}, $\mathnormal{l}_{ac}^{(t)}(\theta, w)$ is the area constraints loss, and $\eta_t$ is a hyper-parameter specified manually to balance the two terms.
Then, the objective loss function is still defined as the accumulation of the loss generated from each side-output layer, which is described in equation \eqref{eq}.



\textbf{Fusion model.}
In order to adequately leverage the multi-scale predictions across all the layers, we merge the side-output layers with each other to generate a fusion layer. $\widehat{Y}_{side}^{(t)}$ is the predicted probability map at the $t$th side output layer.
The output of the fusion layer is defined as 
\begin{equation}
\widehat{Y}_{fuse}=\sum_{t=1}^T\alpha_{t} \widehat{Y}_{side}^{(t)},
\end{equation}
where $\alpha_t$ refers to the weight deduced for the probability map generated by the $t$th side-output layer.

Then, the fusion loss function is given as:
\begin{equation}
\mathcal{L}_{fuse}{(\theta, w)}=  \mathnormal{l}_{mil}^{(fuse)}(\theta, w) + \eta_{fuse}\cdot\mathnormal{l}_{ac}^{(fuse)}(\theta, w),
\end{equation}
where $\mathnormal{l}_{mil}^{(fuse)}(\theta, w)$ is the MIL loss of $\widehat{Y}_{fuse}$ computed as equation \eqref{eq:dws_mil_side}, $\mathnormal{l}_{ac}^{(fuse)}(\theta, w)$ is the area constraints loss of $\widehat{Y}_{fuse}$ computed as equation \eqref{eq:ac_loss}, and $\eta_{fuse}$ is a hyper-parameter specified manually to balance the two terms.
The final objective loss function is defined as below:
\begin{equation}  \mathcal{L}(\theta, w)=\mathcal{L}_{side}(\theta, w)+\mathcal{L}_{fuse}(\theta, w).                     \end{equation}
In the end, we minimize the overall loss function by stochastic gradient descent algorithm during network training:
\begin{equation} {(\theta, w)}^{*}=argmin_{\theta, w} \mathcal{L}(\theta, w).            
\label{eq:overall}
\end{equation}

To summarize, equation \eqref{eq:overall} gives the overall function to learn, which is under the general multiple instance learning with an end-to-end learning process.
Our algorithm is built on top of fully convolutional networks with deep weak supervision and additional area constraints.
The pipeline of our algorithm is illustrated in Figure \ref{fig:flowchart_DWS}.
In our framework, we adopt the first three stages of the VGGNet and then the last convolutional layer of each stage is connected to side-output.
Pixel-level prediction maps can be produced by each side-output layer and the fusion layer.
The fusion layer takes a weighted average of all side-outputs.
The MIL formulation guides the learning of the entire network to
make pixel-level prediction for a better prediction of the image-level labels via softmax functions. 
In each side-output layer, the loss function $l_{mil}$ is computed in the form of deep weak supervision.
Furthermore, area constraints loss $l_{ac}$ makes it possible to constrain the size of predicted cancerous tissues.
Finally, the parameters of our network is learned by minimizing the objective function defined in equation \eqref{eq:overall} via back-propagation using the stochastic gradient descent algorithm.

\subsection{Super-pixels}
Treating each pixel as an instance may sometimes produce jagged tissue boundaries. We therefore alternatively explore another option of defining instances, super-pixels.
Using super-pixels gives rise to a smaller number of instances and consistent elements that can be readily pre-computed using an over-segmentation algorithm \cite{achanta2012slic}.
Our effort starts with the SLIC method mentioned in \cite{achanta2012slic} to generate super-pixels by  grouping the input image pixels into a number of small regions. These super-pixels act as our instances but our main formulation stays the same as to minimize the overall objective function defined in equation \eqref{eq:overall}.

\section{Network Architecture}

We choose the 16-layer VGGNet \cite{simonyan2014very} as the CNN architecture of our framework, which was pre-trained on the ImageNet 1K class dataset and achieved state-of-the-art performance in the ImageNet challenge \cite{ILSVRCarxiv14}.
Although ImageNet consists of natural images, which are different from histopathology images, several previous works \cite{xu2015deep} have shown that networks pre-trained on  ImageNet are also very effective in dealing with histopathology images.
The VGGNet has 5 sequential stages before the fully-connected layer. Within each stage, two or three convolutional layers are followed by a 2-stride pooling layer.
In our framework, we trim off the 4th and 5th stages and only adopt the first three stages.
Side-output layers are connected to last convolutional layer in each stage (see Table \ref{table:results_vgg}).
The side-output layer is a $1\times 1$ convolutional layer of one-channel output with the sigmoid activation.
This style of network architecture makes different side-output layers have different strides and receptive field sizes, resulting in side-outputs of different scales.
Having three side-output layers, we add a fusion layer that takes a weighted average of side-outputs to yield the final output.
Note due to the different strides in different side-output layers, the sizes of different side-outputs are not same.
Hence, before the fusion, all side-outputs are upsampled to the size of the input image by bilinear interpolation.

\begin{table}[!ht]
\renewcommand{\arraystretch}{1}
\caption{The receptive field size and stride in the VGGNet \cite{simonyan2014very}. In our framework, the first three stages are used, and the bolded parts indicate convolutional layers linked to additional side-output layers.}
\label{table:results_vgg}
\centering
\begin{tabular}{c|ccccc}
\hline
layer &\bf{c1\_2} &\bf{c2\_2} &\bf{c3\_3} &c4\_3 &c5\_3 \\
\hline
rf size &\bf{5} &\bf{14} &\bf{40} & 92 & 196 \\
\hline
stride &\bf{1} &\bf{2} &\bf{4} & 8 & 16 \\
\hline
\end{tabular}
\end{table}

\begin{figure}[!ht]
\centering
\subfloat[input]{\includegraphics[width=0.12\linewidth]{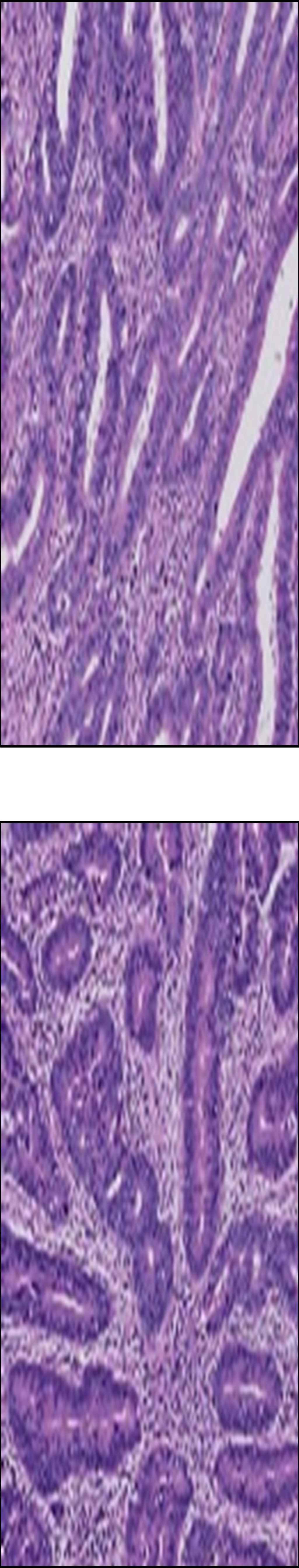}}
\hspace{0.004\linewidth}
\subfloat[side1]{\includegraphics[width=0.12\linewidth]{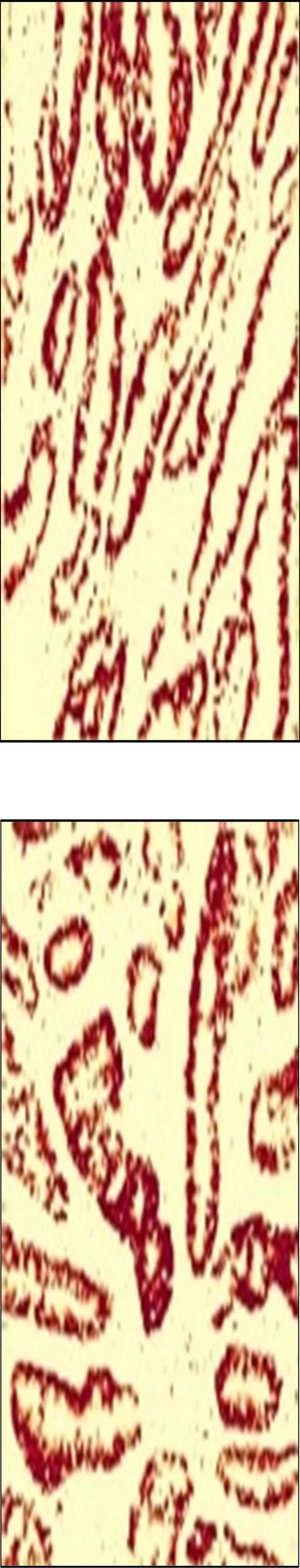}}
\hspace{0.004\linewidth}
\subfloat[side2]{\includegraphics[width=0.12\linewidth]{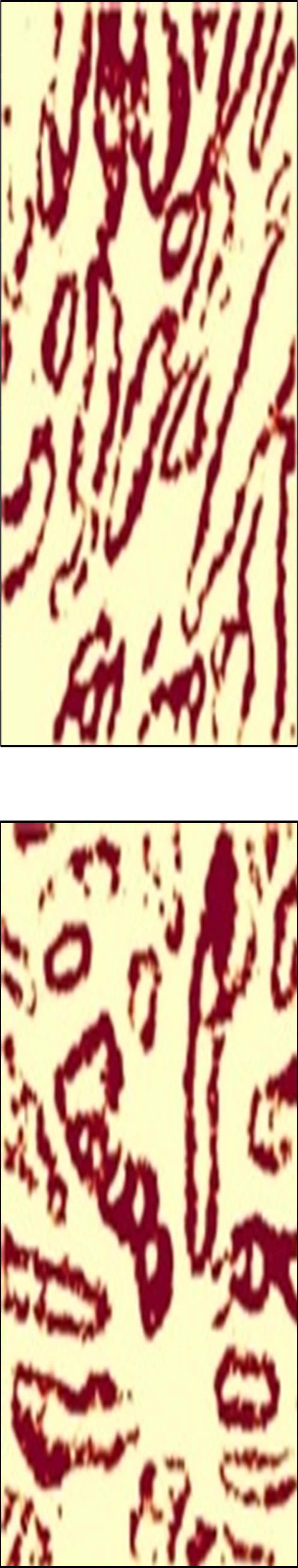}}
\hspace{0.004\linewidth}
\subfloat[side3]{\includegraphics[width=0.12\linewidth]{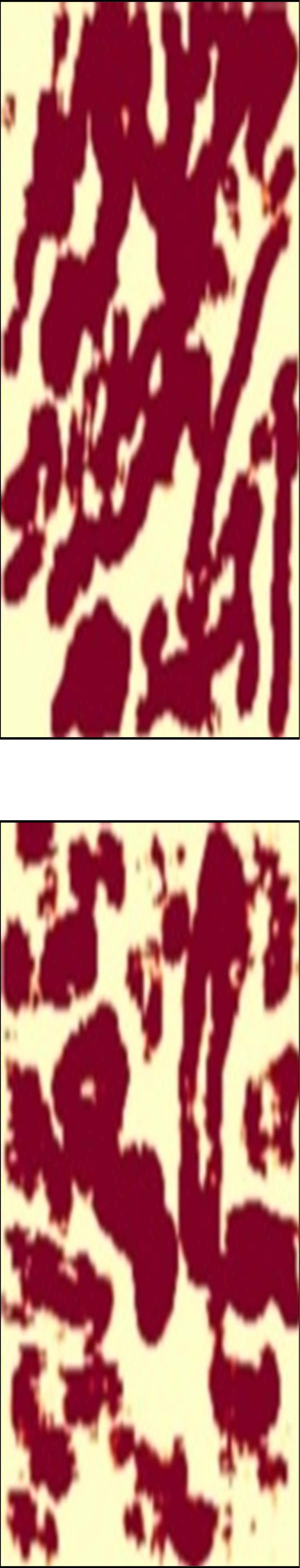}}
\hspace{0.004\linewidth}
\subfloat[side4]{\includegraphics[width=0.12\linewidth]{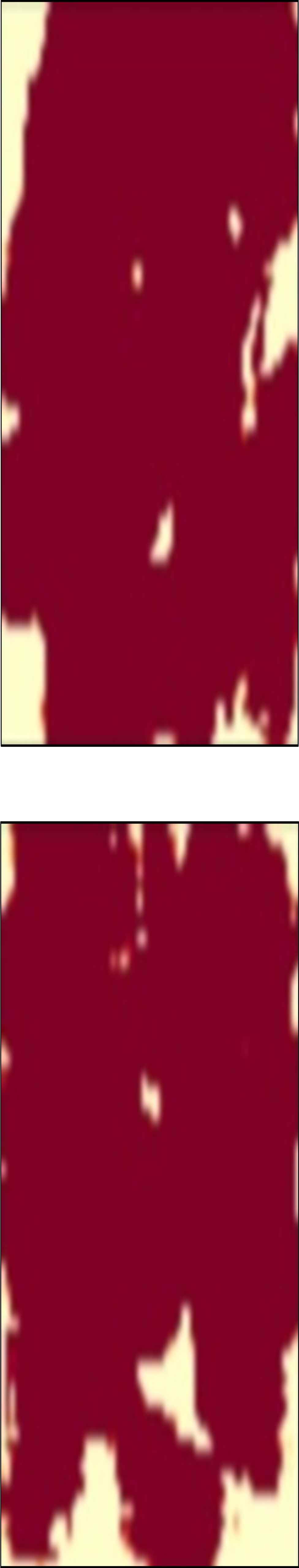}}
\hspace{0.004\linewidth}
\subfloat[side5]{\includegraphics[width=0.12\linewidth]{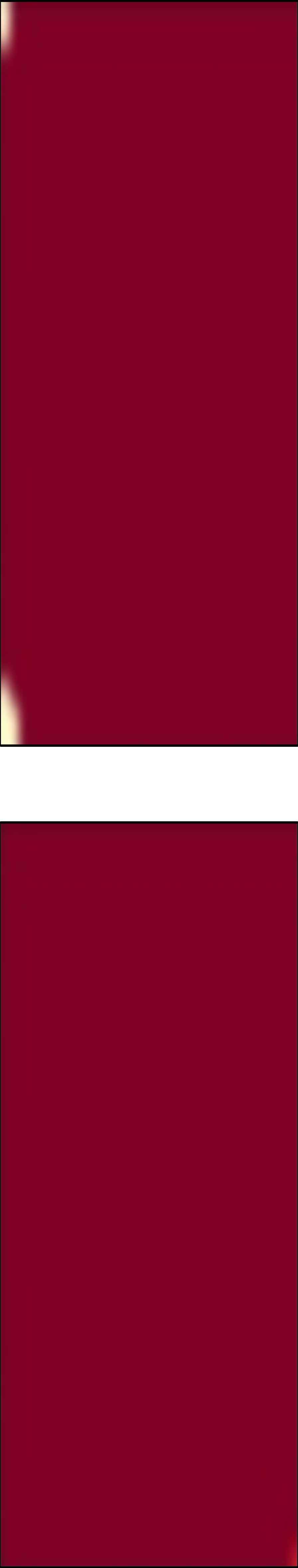}}
\hspace{0.004\linewidth}
\caption{Side-outputs from 5 stages of the VGGNet. As the network going deeper, the receptive field size increases and the side-output grows to be larger and coarser.
In the 4th and 5th stages, almost all the pixels are recognized as positive, and then positive areas almost cover the entire images. Therefore, we trim off the 4th and 5th stages in our framework.}
\label{fig:concept_sideout}
\end{figure}

\textbf{The reason for trimming the VGGNet.}
In histopathology images, tissues appear as local texture patterns.
In the 4th and 5th stages of the VGGNet, the receptive field sizes (see Table \ref{table:results_vgg}) become too large for local textures.
Figure \ref{fig:concept_sideout} shows side-outputs if all 5 stages of the VGGNet is adopted.
From the figure, as the network going deeper, the receptive field size increases and the side output grows to be larger and coarser. In the 4th and 5th stages, the side-outputs almost fill the entire images, which becomes meaningless.
Thus we ignore the 4th and 5th stages of the VGGNet in our framework, due to their overlarge receptive field size.

\section{Experiments}
\label{Experiments}

In this section, we first describe the implementation details of our framework.
Two histopathology image datasets are used to evaluate our proposed methods.

\subsection{Implementation}

We implement our framework on top of the publicly available Caffe toolbox \cite{jia2014caffe}.
Based on the official version of Caffe, we add a layer to compute the softmax of the generalized mean for pixel-level predictions and a layer to compute the area constraints loss from pixel-level predictions.

\textbf{Model parameters.} The MIL loss is known to be hard to train, and special care is required for choosing training hyper-parameters.
In order to reduce fluctuations in optimizing the MIL loss, all training data are used in each iteration (the mini-batch size is equal to the size of the training set).
The network is trained with Adam optimizer \cite{kingma2014adam}, using a momentum of $0.9$, a weight decay of $0.0005$, and a fixed learning rate of $0.001$.
The learning rates of side-output layers are set to $1/100$ of the global learning rate. 
For the parameter of the generalized mean, we set $r=4$. 

\textbf{Fusion layer.} The fusion layer adopts the weighted average of side-output layers.
At the first attempt, we initialize all the fusion weights to $1/3$, and let the model learn appropriate weights in the training phase.
When the network converges, we observe that the outputs of the fusion layer are very close to the 3rd side-output layer, making the fusion results useless.
The reason for this outcome is that for the deeper side-output layer, it has a lower MIL loss as a result of more discriminative features.
To resolve the problem, we use fixed fusion weights instead of learning them.
Based on cross-validation on training data, the fusion weights are finally chosen as $0.2,0.35,0.45$ for the three side-output layers, and a threshold of $0.5$ is used to produce segmentation results.

\textbf{Weight of area constraints loss.} The weight of the area constraints loss is crucial for CDWS-MIL, since it directly decides the strength of constraints.
Strong constraints may make the network unable to converge, while weak constraints have a little help with learning better segments.
To decide the appropriate loss weight, we select a validation set from training data to evaluate different options.
The loss weights of area constraints for the different side-output layers are decided separately.
To achieve this, when deciding the loss weight for a side-output layer, only this layer has area constraints.
Finally, loss weights of $2.5,5,10,10$ are selected for the three side-output layers and the fusion layer.

\subsection{Experiment A}

\textit{Dataset A} is a histopathology image dataset of colon cancer, which consists of 330 cancer (CA) and 580 non-cancer (NC) images.
In this dataset, 250 cancer and 500 non-cancer images are used for training; 80 cancer and 80 non-cancer images are used for testing.
These images are obtained from the NanoZoomer 2.0HT digital slide scanner produced by Hamamatsu Photonics with a magnification factor of 40,
i.e. 226 nm/pixel. Each image has a resolution of $3,000\times 3,000$.
Two pathologists were asked to label each image to be cancerous or non-cancerous.
When the two pathologists disagree on a particular image, they would discuss it with another senior pathologist to reach an agreement.
For the evaluation purpose, we also ask the pathologists to annotate all the cancerous tissues for each image, which are only used in testing to evaluate our algorithms. For simplicity, in our table we use CA to refer to cancer images and use NC to refer to  non-cancer images.

F-measure is used as the evaluation metric for experiments on \textit{Dataset A}.
Given the ground truth map $G$ and the prediction map $H$,
we define $\text{F-measure}=(2 \cdot \text{precision} \times \text{recall})/(\text{precision}+\text{recall})$ in which $\text{precision}=|H\cap G|/|H|$ and $\text{recall}=|H\cap G|/|G|$.
For images with label $Y=1$, the prediction map consists of pixels with $1$ as the pixel-level prediction, and the ground truth map is the annotated cancerous regions.
For images with label $Y=0$, the prediction map consists of pixels with $0$ as the pixel-level prediction, and the ground truth map is the entire image.

\begin{figure*}[!htp]
\centering
\includegraphics[width=150mm]{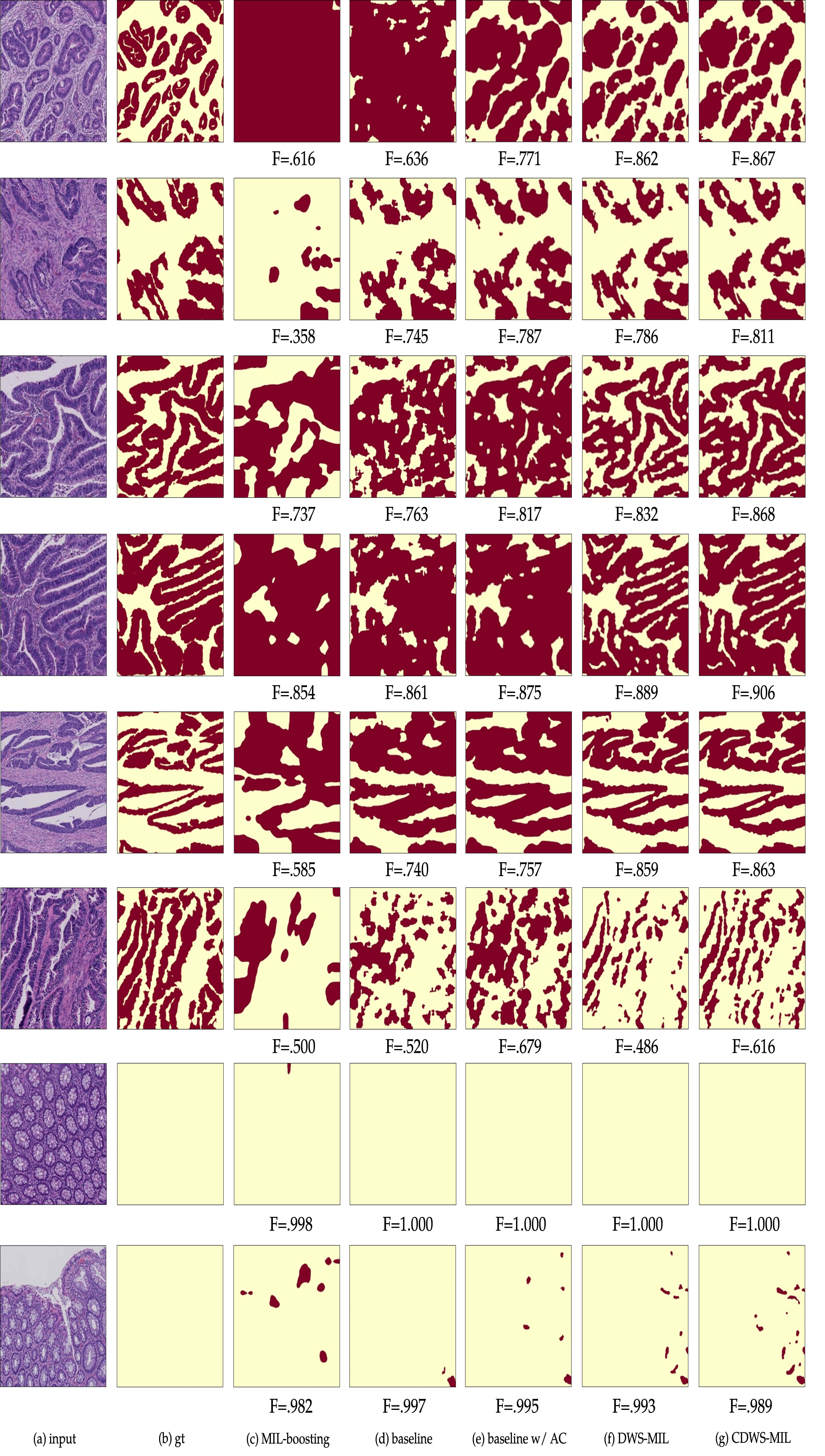}
\caption{Segmentation results on \textit{dataset A}:
(a) Input images.
(b) Ground truth labels.
(c) Results by MIL-Boosting.
(d) Results by our baseline.
(e) Results by our baseline w/ AC.
(f) Results by DWS-MIL.
(g) Results by CDWS-MIL.
Compared with MIL-Boosting (patch-based), our proposed DWS-MIL and CDWS-MIL produce significantly improved results due to the characteristics we introduced in this paper.   
}
\label{fig:resultsA_all_training_data}
\end{figure*}

\textbf{Comparisons.} Table \ref{table:results_all_training_data} summarizes the results of our proposed algorithms and other methods on \textit{Dataset A}.
In all experiments, images are resized to $500\times 500$ pixels for time-efficiency.
In MIL-Boosting, a patch size of $64 \times 64$ pixels and a stride of 4 pixels are used for both training and testing, and other settings follow \cite{xu2014weakly}.
To show the effectiveness of area constraints, we also integrate area constraints into our baseline, denoted as ``our baseline w/ AC'' in the table.
From the table, DWS-MIL and CDWS-MIL surpass other methods by large margins,
and constrained deep weak supervision contributes an improvement of 7.3\% than our baseline method (0.835 vs 0.778).
Figure \ref{fig:resultsA_all_training_data} shows some examples of segmentation results by these methods.

\begin{table}[!ht]
\renewcommand{\arraystretch}{1.2}
\caption{Performance of various methods on \textit{Dataset A}.}
\label{table:results_all_training_data}
\centering
\begin{tabular}{l|c|c}
\hline
Method & F-measure of CA & F-measure of NC \\
\hline
MIL-Boosting & 0.684 & 0.997 \\
our baseline & 0.778 & 0.998 \\
our baseline w/ AC & 0.815 & 0.998 \\
DWS-MIL & 0.817 & 0.999 \\
CDWS-MIL & \bf{0.835} & 0.997 \\
\hline
\end{tabular}
\end{table}

\begin{figure}[!htp]
\centering
\subfloat[input]{\includegraphics[width=0.13\linewidth]{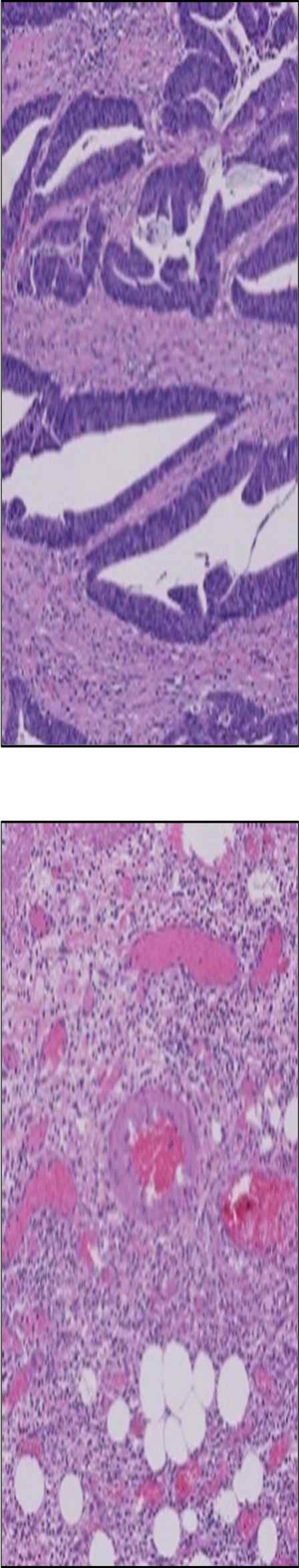}}
\hspace{0.002\linewidth}
\subfloat[gt]{\includegraphics[width=0.13\linewidth]{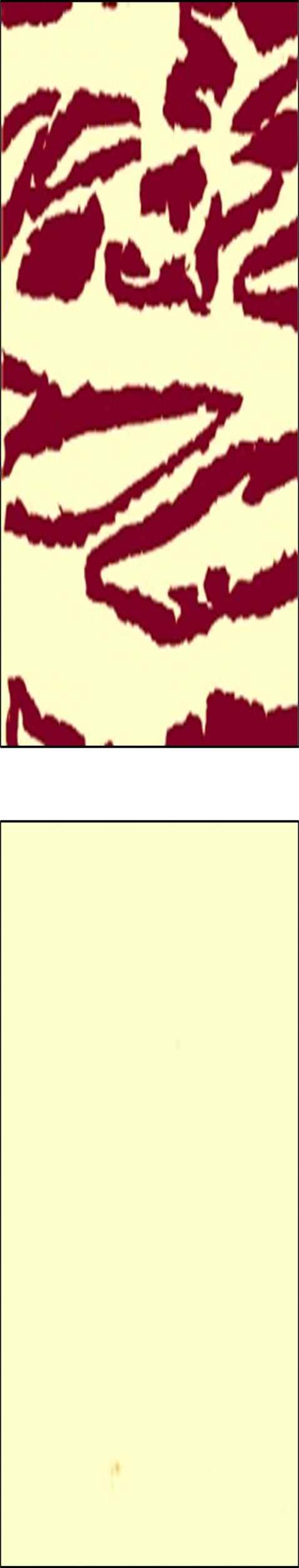}}
\hspace{0.002\linewidth}
\subfloat[20\%]{\includegraphics[width=0.13\linewidth]{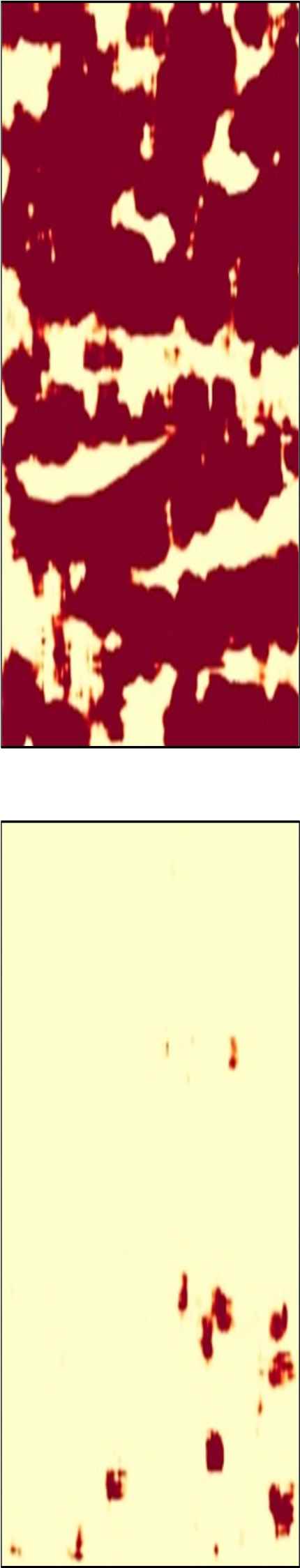}}
\hspace{0.002\linewidth}
\subfloat[40\%]{\includegraphics[width=0.13\linewidth]{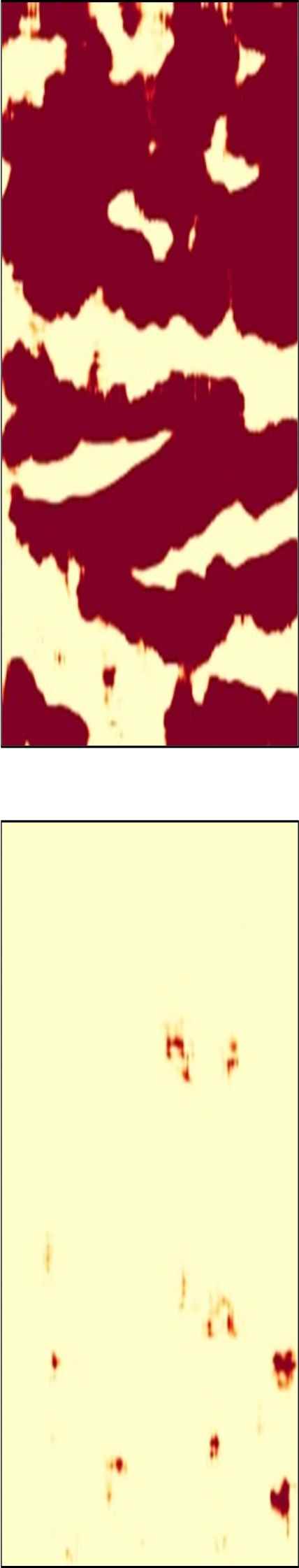}}
\hspace{0.002\linewidth}
\subfloat[60\%]{\includegraphics[width=0.13\linewidth]{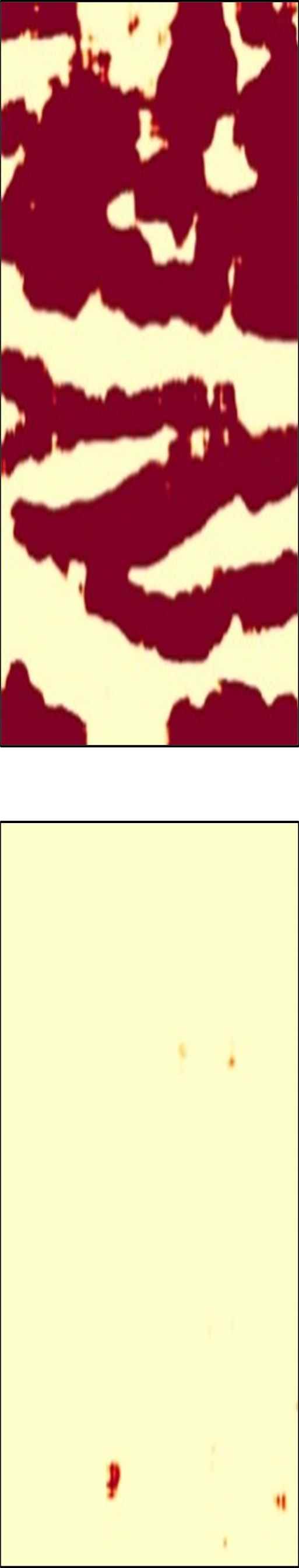}}
\hspace{0.002\linewidth}
\subfloat[80\%]{\includegraphics[width=0.13\linewidth]{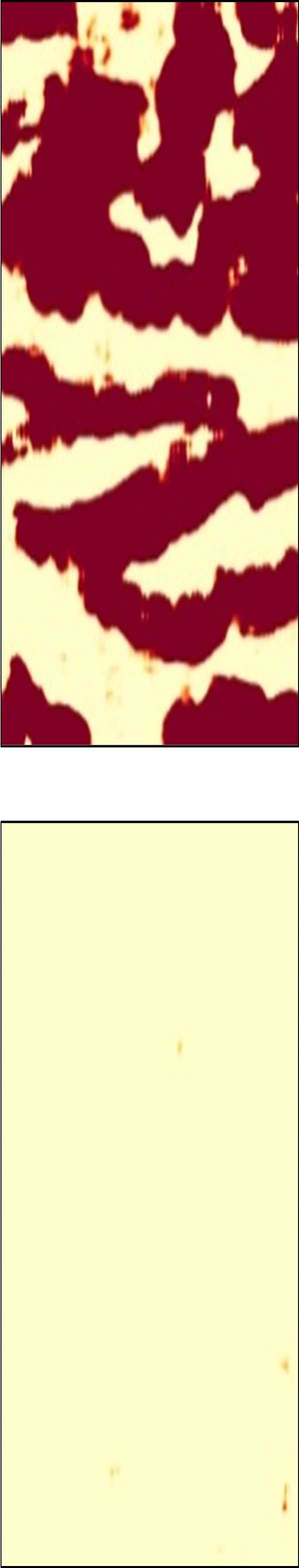}}
\hspace{0.002\linewidth}
\subfloat[100\%]{\includegraphics[width=0.13\linewidth]{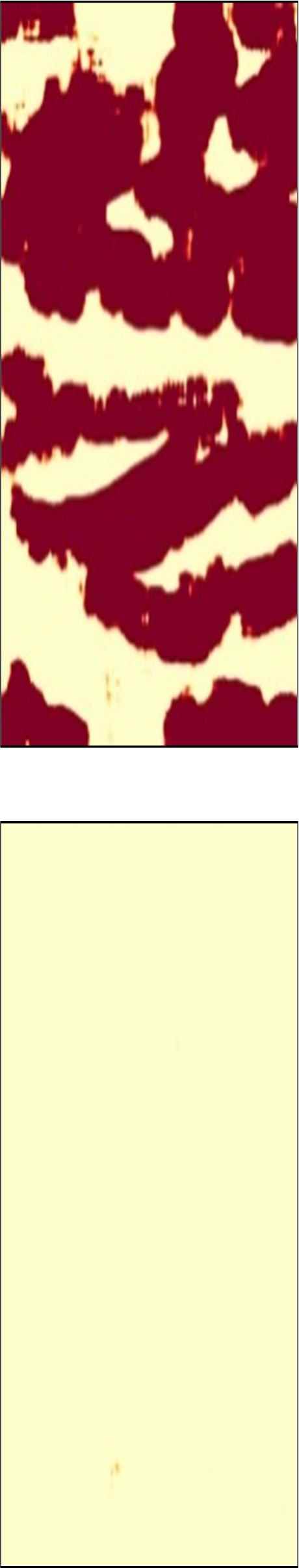}}
\hspace{0.002\linewidth}
\caption{Differences in results with different amounts of training data:
(a) The input images.
(b) Ground truth labels.
(c) Results that use 20\% of training data.
(d) Results that use 40\% of training data.
(e) Results that use 60\% of training data.
(f) Results that use 80\% of training data.
(g) Results that use all the training data.
}
\label{fig:results_different_training}
\end{figure}

\textbf{Less training data.} To observe how the amounts of training data influence our baseline method, we train our baseline with less training data.
Table \ref{table:results_different_training} summarizes the results, and Figure \ref{fig:results_different_training} shows some samples of segmentation results that use different amounts of training data.
Given more training data, the performance of segmentation is better. In the case of less training data, the segmentation results tend to be larger than the ground truth.
This observation can be explained by analyzing the MIL formulation.
From the expression of the MIL loss, identifying more pixels as positive in a positive image always results in a lower MIL loss.
With a smaller amount of negative training images, it is easier to achieve this objective.

\begin{table}[!ht]
\renewcommand{\arraystretch}{1.2}
\caption{Performance of our baseline trained with less training data.}
\label{table:results_different_training}
\centering
\begin{tabular}{l|c|c|c|c}
\hline
{Training data} &
\multicolumn{2}{c|}{F-measure of CA} &
\multicolumn{2}{c}{F-measure of NC} \\
\cline{2-5}
(Pos,Neg)&w/o AC&w/ AC&w/o AC&w/ AC\\
\hline
20\% (50,100)&0.758&0.801&0.997&0.997\\
40\% (100,200)&0.762&0.809&0.997&0.999\\
60\% (150,300)&0.778&0.805&0.999&0.999\\
80\% (200,400)&0.779&0.813&0.998&0.999\\
100\% (250,500)&0.778&0.815&0.998&0.998\\
\hline
\end{tabular}
\end{table}

\begin{figure}[!htp]
\centering
\subfloat[input]{\includegraphics[width=0.13\linewidth]{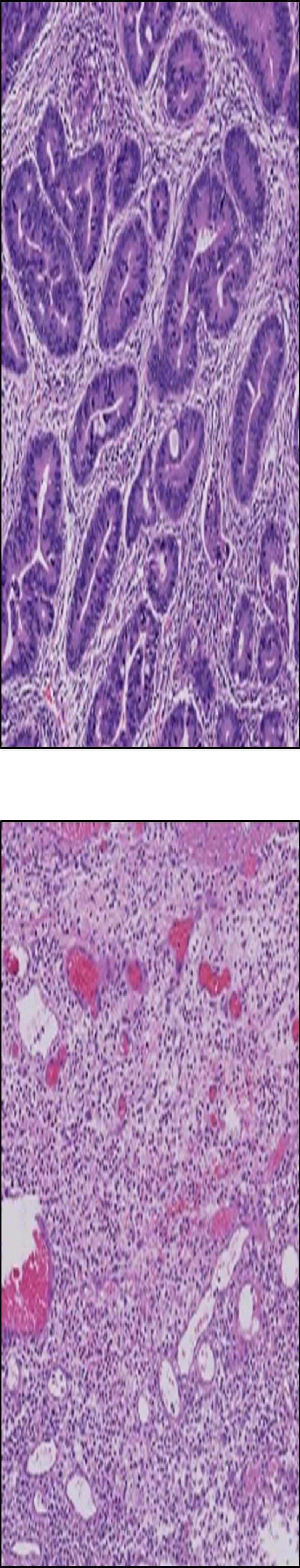}}
\hspace{0.002\linewidth}
\subfloat[gt]{\includegraphics[width=0.13\linewidth]{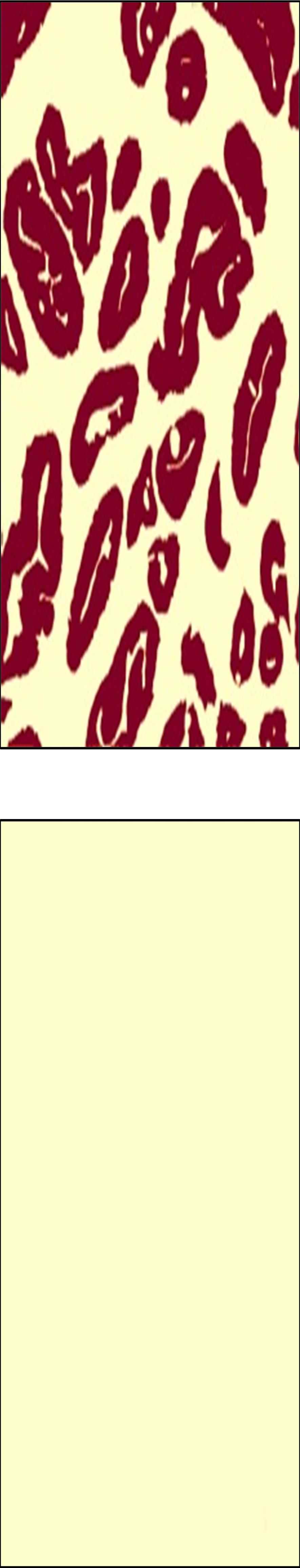}}
\hspace{0.002\linewidth}
\subfloat[w/o AC]{\includegraphics[width=0.13\linewidth]{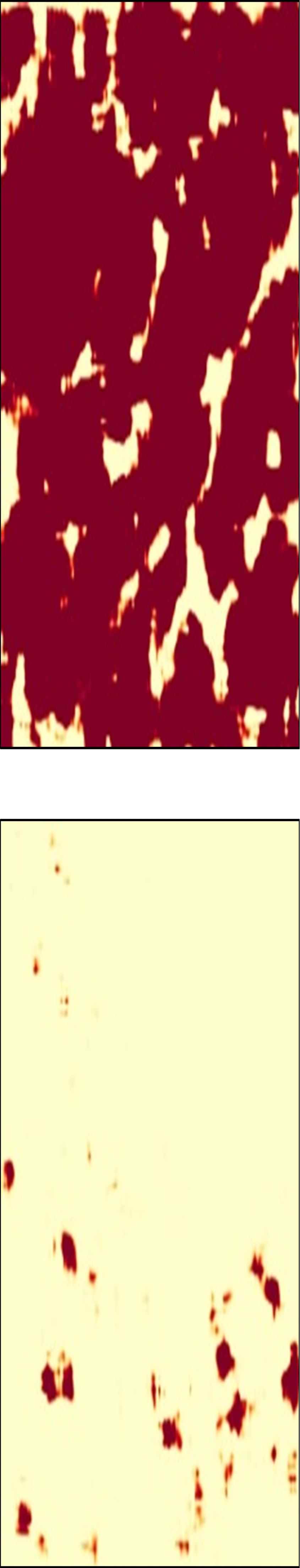}}
\hspace{0.002\linewidth}
\subfloat[w/ AC]{\includegraphics[width=0.13\linewidth]{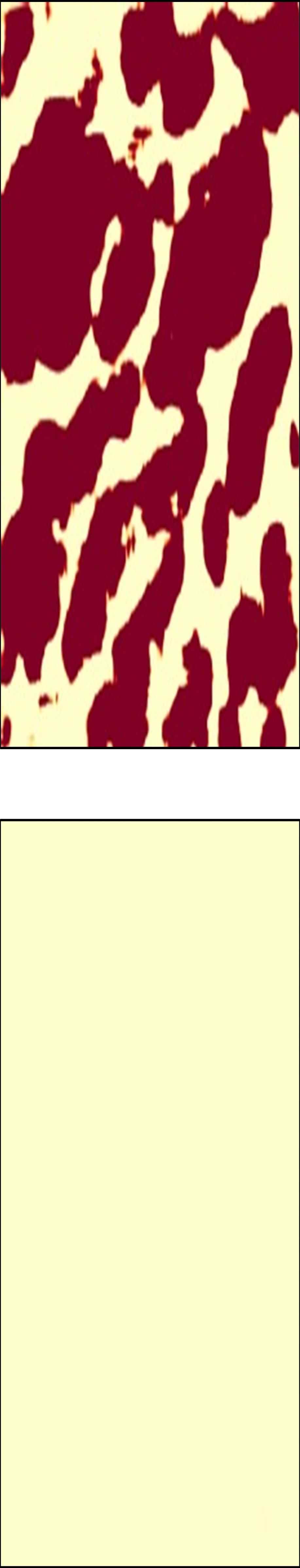}}
\hspace{0.002\linewidth}
\caption{Comparison of using and not using area constraints:
(a) The input images.
(b) Ground truth labels.
(c) Results of our baseline.
(d) Results of our baseline w/ AC.
The area constraints loss constrains the model to learn better segmentations.
}
\label{fig:results_area_constraints}
\end{figure}

\textbf{Area constraints.} From Table \ref{table:results_different_training}, the area constraints enable our baseline method to achieve a competitive accuracy with a small training set.
Equipped with area constraints, our baseline method  using 20\% of training data achieves better accuracy than using all training data without area constraints.
Figure \ref{fig:results_area_constraints} shows some samples of segmentation results by using and not using area constraints.
It is clear that area constraints achieve the goal of constraining the model to learn smaller segmentations,
which significantly improves segmentation accuracy for both cancer images and non-cancer images.
When not using area constraints, the segmentation results are much larger than the ground truth, and also have the tendency to cover entire images.
In contrast, when the area constraints loss is integrated with the MIL loss, the fact that too many pixels are identified as positive will yield a large area constraint loss to compete with the MIL loss.
To achieve a balance between the MIL loss and the area constraints loss, it only learns the most confident pixels as positive, as proven in Figure \ref{fig:results_area_constraints}.
Table \ref{table:results_partial_training_data} summarizes results of the baseline methods, DWS-MIL, CDWS-MIL and MIL-Boosting using 20\% of training data.
Comparing CDWS-MIL in Table \ref{table:results_partial_training_data} with other methods in Table \ref{table:results_all_training_data},
CDWS-MIL outperforms other methods using only 20\% of training data.
In addition, constrained deep weak supervision contributes an improvement of 8.2\% over our baseline method (0.820 vs 0.758), which is larger than the condition that all training data are used.

\begin{table}[!ht]
\renewcommand{\arraystretch}{1.2}
\caption{Performance of various methods with 20\% training data.}
\label{table:results_partial_training_data}
\centering
\begin{tabular}{l|c|c}
\hline
Method &F-measure of CA& F-measure of NC \\
\hline
MIL-Boosting & 0.635 & 0.995 \\
our baseline& 0.758 & 0.997 \\
our baseline w/ AC& 0.800 & 0.997 \\
DWS-MIL& 0.808 & 0.998 \\
CDWS-MIL& \bf{0.820} & 0.998 \\
\hline
\end{tabular}
\end{table}

\begin{figure}[!htp]
\centering
\subfloat[input]{\includegraphics[width=0.13\linewidth]{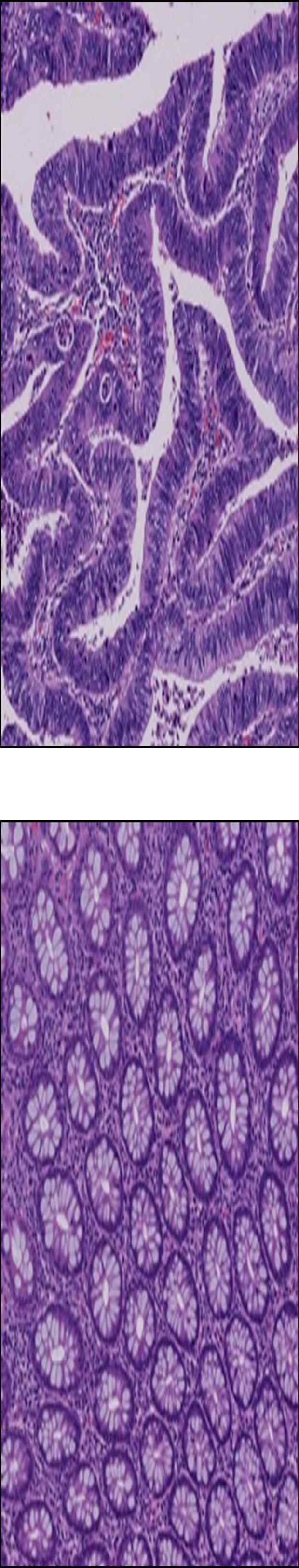}}
\hspace{0.002\linewidth}
\subfloat[gt]{\includegraphics[width=0.13\linewidth]{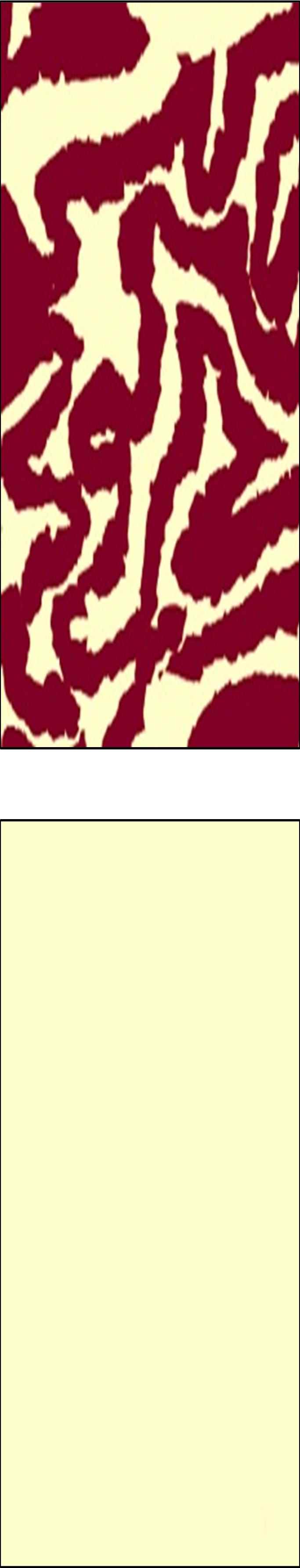}}
\hspace{0.002\linewidth}
\subfloat[side1]{\includegraphics[width=0.13\linewidth]{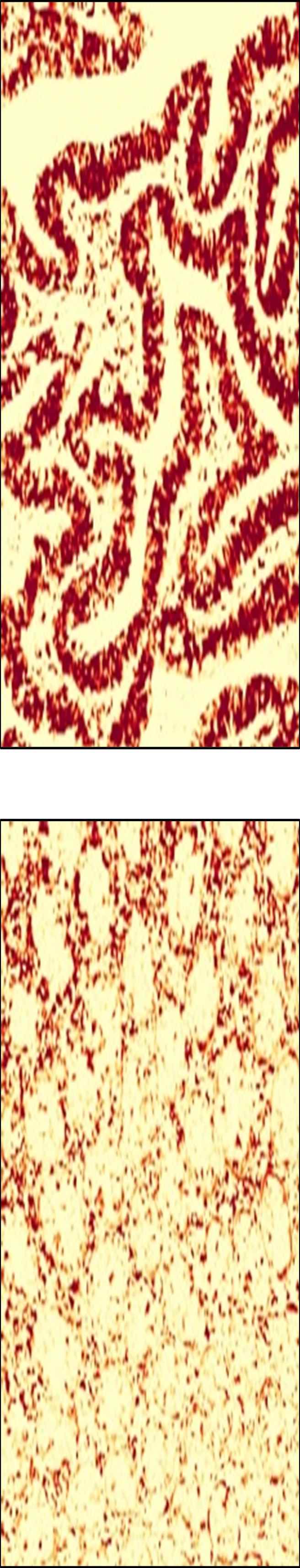}}
\hspace{0.002\linewidth}
\subfloat[side2]{\includegraphics[width=0.13\linewidth]{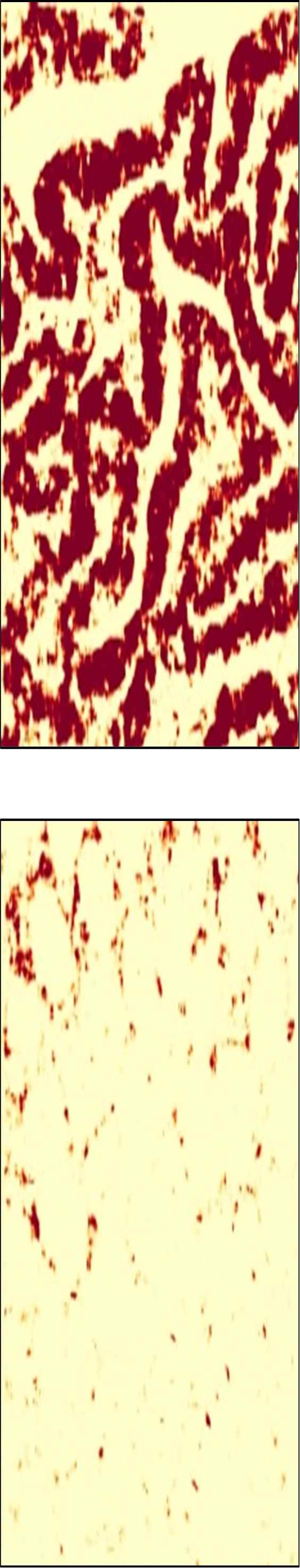}}
\hspace{0.002\linewidth}
\subfloat[side3]{\includegraphics[width=0.13\linewidth]{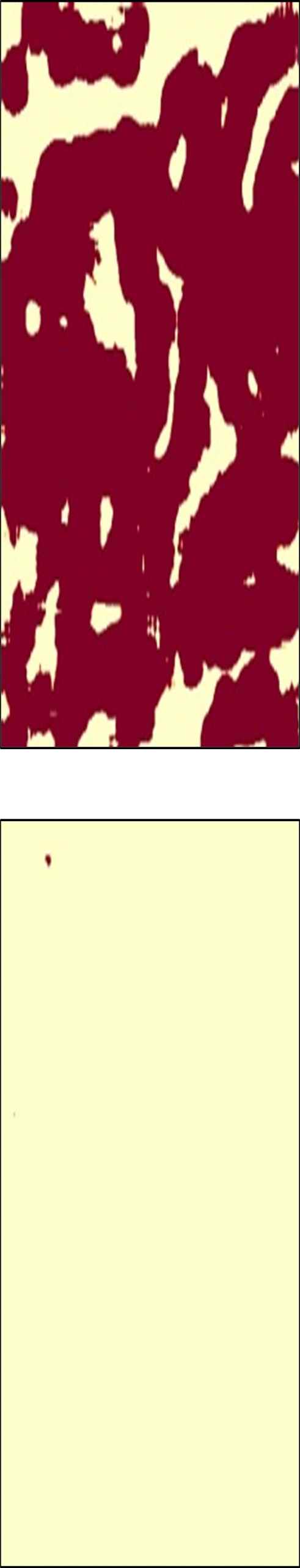}}
\hspace{0.002\linewidth}
\subfloat[fusion]{\includegraphics[width=0.13\linewidth]{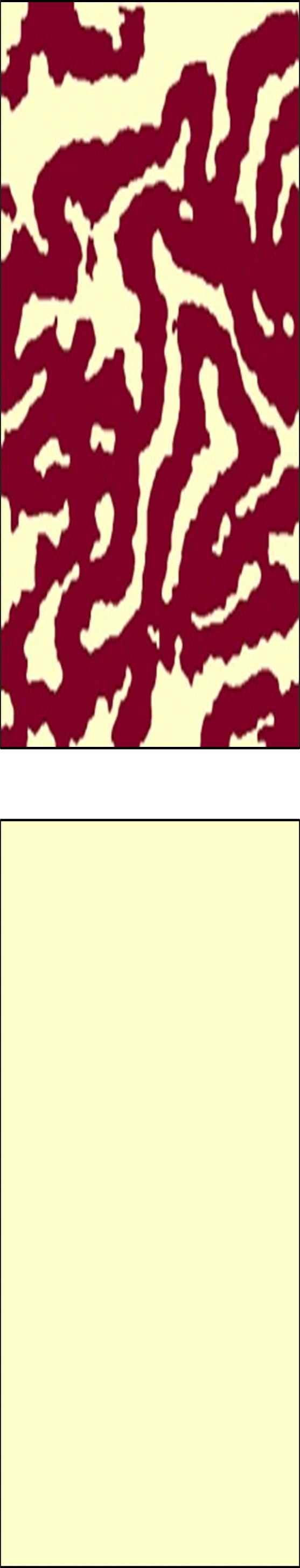}}
\hspace{0.002\linewidth}
\caption{Results of side-output layers:
(a) The input images.
(b) Ground truth labels.
(c) Results of side-output 1.
(d) Results of side-output 2.
(e) Results of side-output 3.
(f) Results by final fusion.
The figure shows a nesting characteristic of segmentation outputs from the lower side-output layer to the higher side-output layer. The final fusion balances pros and cons of different side outputs, and achieves better segmentation results than all of them.
}
\label{fig:results_dsn_layers}
\end{figure}

\textbf{Deep weak supervision.} To illustrate the effectiveness of deep weak supervision, Table \ref{table:results_dsn_layers} summarizes segmentation accuracies of the different side-outputs and Figure \ref{fig:results_dsn_layers} shows some examples of the different side-outputs. From Table \ref{table:results_dsn_layers}, we observe that segmentation accuracy improves from lower layers to higher ones. Figure \ref{fig:results_dsn_layers} shows pixel-level predictions (segmentation) of side-output layer 1, side-output layer 2, and side-output layer 3. This is understandable since the receptive fields of CNN become increasingly bigger from lower layers to higher ones. Histopathology images typically observe local texture patterns. The final fusion layer that combines all the intermediate layers achieves the best result.

\begin{table}[ht]
\renewcommand{\arraystretch}{1.2}
\caption{Performance of different side-output layers. The first line: DWS-MIL; The second line: CDWS-MIL.}
\label{table:results_dsn_layers}
\centering
\begin{tabular}{m{0.07\linewidth}|m{0.07\linewidth}|m{0.07\linewidth}|m{0.07\linewidth}|m{0.07\linewidth}|m{0.07\linewidth}|m{0.07\linewidth}|m{0.07\linewidth}}
\hline
\multicolumn{4}{c|}{F-measure of CA}&
 \multicolumn{4}{c}{F-measure of NC}\\
\cline{1-8}
 side1&side2&side3&fusion&side1&side2&side3&fusion \\
\hline
0.666 & 0.747 & 0.783 & 0.817 & 0.984 & 0.994 & 0.997 & 0.999 \\
0.660 & 0.783 & 0.819 & 0.835 & 0.984 & 0.994 &0.997 & 0.997 \\
\hline
\end{tabular}
\end{table}

\begin{figure}[!htp]
\centering
\subfloat[]{\includegraphics[width=0.13\linewidth]{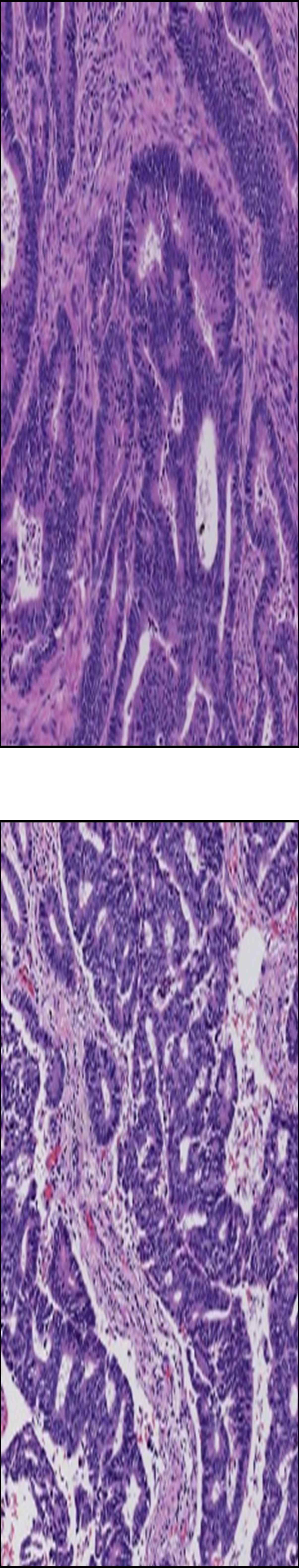}}
\hspace{0.002\linewidth}
\subfloat[]{\includegraphics[width=0.13\linewidth]{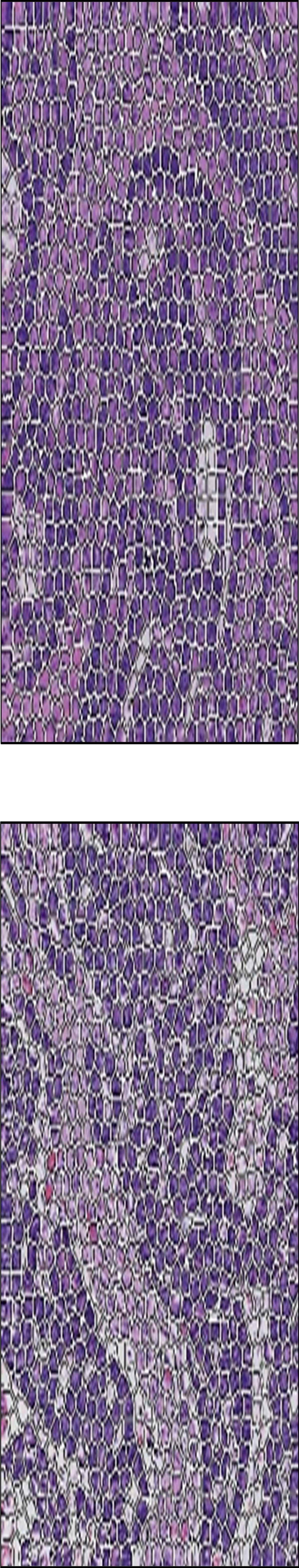}}
\hspace{0.002\linewidth}
\subfloat[]{\includegraphics[width=0.13\linewidth]{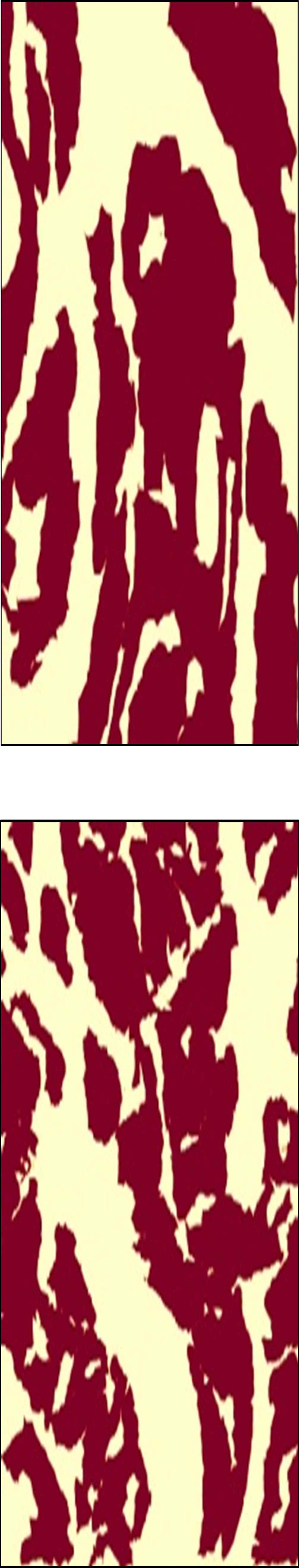}}
\hspace{0.002\linewidth}
\subfloat[]{\includegraphics[width=0.13\linewidth]{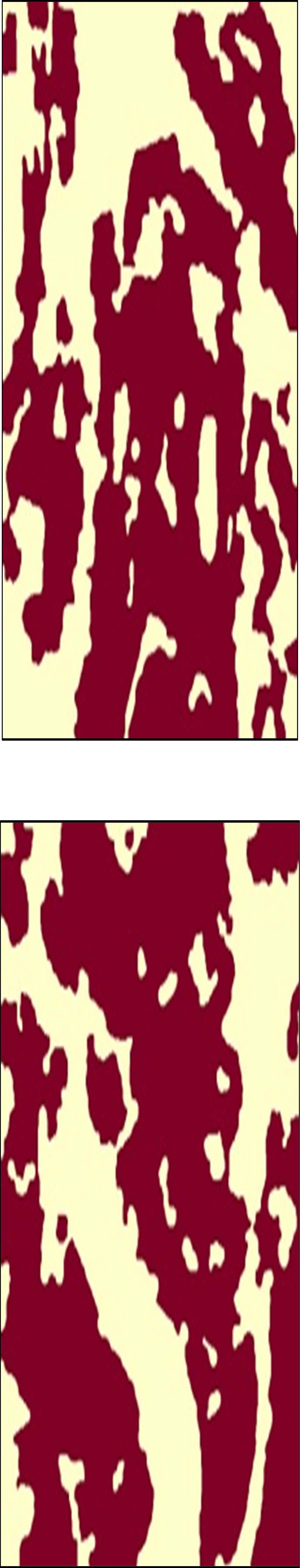}}
\hspace{0.002\linewidth}
\subfloat[]{\includegraphics[width=0.13\linewidth]{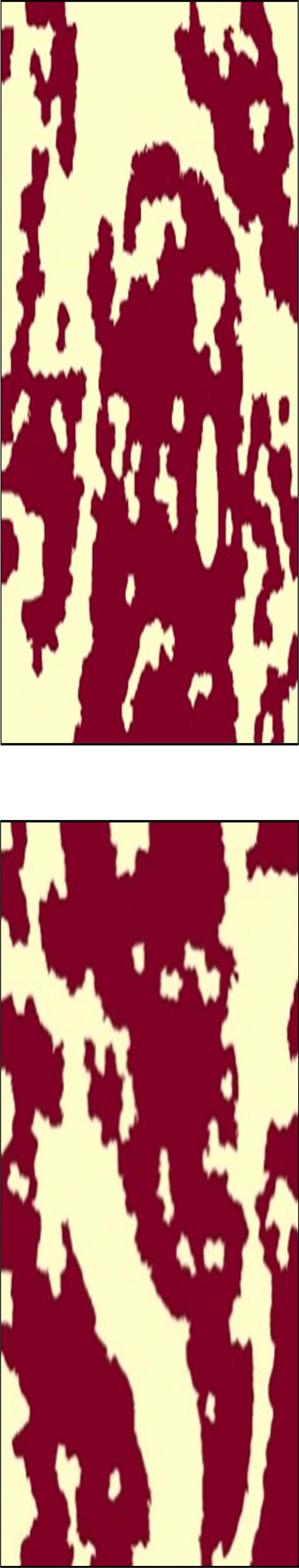}}
\hspace{0.002\linewidth}
\caption{Comparisons of DWS-MIL and DWS-MIL w/ super-pixel:
(a) The input images.
(b) Results generated by SLIC method \cite{achanta2012slic}.
(c) Ground truth labels.
(d) Results of DWS-MIL.
(e) Results of DWS-MIL w/ super-pixel.
Some detailed edges can be recognized with the help of super-pixels. 
}
\label{fig:result_superpixel.jpg}
\end{figure}

\textbf{Super-pixels.} We conduct experiments to compare DSW-MIL and DSW-MIL w/ super-pixel.
We adopt the SLIC method \cite{achanta2012slic} to generate super-pixels. The average F-measures of DSW-MIL w/ super-pixel on cancer images and non-cancer images are 0.818 and 0.999, respectively.
Figure \ref{fig:result_superpixel.jpg} shows some samples of the segmentation results of the two methods.
In histopathology images, super-pixels adhere well to tissue edges, resulting in more accurate segmentations.
The adoption of super-pixels can help to predict more detailed boundaries.


\textbf{Advantages of CDWS-MIL.} MIL-Boosting in comparison is a patch-based MIL approach.
The bags in their MIL formulation are composed of patches sampled from input images.
Figure \ref{fig:resultsA_all_training_data} shows some samples of segmentation results of CDWS-MIL and MIL-Boosting, demonstrating that  in some cases (like the 2nd row in the figure), MIL-Boosting completely fails to learn the correct segmentations, while in other cases (like the 5th row in the figure), CDWS-MIL and MIL-Boosting both learn roughly correct segmentations, but CDWS-MIL learns much more elaborate ones.
There are three advantages of our framework CDWS-MIL over MIL-Boosting:
(1) CDWS-MIL is an end-to-end segmentation framework, which can learn more detailed segmentations than the patch-based MIL-Boosting;
(2) Deep weak supervision enables CDWS-MIL to learn from multiple scales, and the fusion output balances outputs of different scales to achieve the best accuracy;
(3) Area constraints in CDWS-MIL are straightforward, while being hard to be integrated into patch-based methods like MIL-Boosting.

\subsection{Experiment B}

\textit{Dataset B} is a histopathology image dataset of 30 colon cancer images and 30 non-cancer images which are referred as tissue microarrays (TMAs).
The dataset is randomly selected from the dataset in \cite{xu2014weakly}. 
All images have a resolution of $1024\times 1024$ pixels, and the rough estimations of the portion of cancerous regions have 8 levels $0.05,0.1,0.15,\ldots,0.4$.
They are annotated in the same way as \textit{Dataset A}.

\begin{figure}[h]
\centering
\includegraphics[width=80mm]{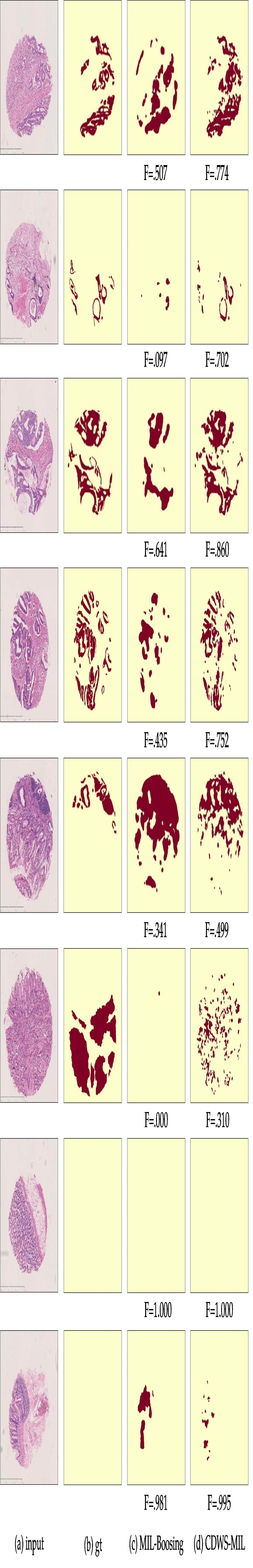}
\caption{Segmentation results on \textit{dataset B}:
(a) Input images.
(b) Ground truth labels.
(c) Results by MIL-Boosting.
(d) Results by CDWS-MIL.
Compared with MIL-Boosting (patch-based), CDWS-MIL produces significantly improved results due to the characteristics we introduced in this paper.   
}
\label{fig:resultsB_all_training_data}
\end{figure}

We conduct experiments to compare MIL-Boosting with our proposed method CDWS-MIL on \textit{Dataset B}.
All experiments are conducted with 5-fold cross-validation, and the evaluation metric is the same on \textit{Dataset A}.
The average F-measures of CDWS-MIL on cancer images and non-cancer images are 0.622 and 0.997, respectively.
The average F-measures of MIL-Boosting on cancer images and non-cancer images are 0.449 and 0.993, respectively.
Figure \ref{fig:resultsB_all_training_data} shows some samples of the segmentation results of these two methods.

\section{Conclusion}
In this paper, we have developed an end-to-end framework under deep weak supervision to perform image-to-image segmentation for histopathology images. To preferably learn multi-scale information, deep weak supervision is developed in our formulation. Area constraints are also introduced in a natural way to seek for additional weakly-supervised information. Experiments demonstrates that our methods attain the state-of-the-art results on large-scale challenging histopathology images. The scope of our proposed methods are quite broad and they can be widely applied to a range of medical imaging and computer vision applications.

\section*{Acknowledgment}
\footnotesize{We would like to thank Lab of Pathology and Pathophysiology, Zhejiang University in China for providing data and help.}

\ifCLASSOPTIONcaptionsoff
  \newpage
\fi



%




\bibliographystyle{IEEEtran}
\bibliography{IEEEabrv,./ref}

%

\end{document}